\documentclass[journal]{IEEEtran}
\usepackage{amsmath,amsthm}
\usepackage{amsmath}
\usepackage{amssymb}
\usepackage{amsfonts}
\usepackage{graphicx}
\graphicspath{{graphics/}}
\usepackage{tikz,tikz-3dplot}
\usepackage{xcolor}
\usepackage{bm}
\usepackage{dsfont}
\usepackage{stfloats}
\usepackage[flushleft]{threeparttable}
\usepackage[numbers,sort&compress]{natbib}

\usepackage[switch]{lineno}

\usepackage[colorlinks]{hyperref}
\hypersetup{colorlinks,breaklinks,linkcolor=blue,urlcolor=blue,anchorcolor=blue,citecolor=blue}
\usepackage{booktabs}
\usepackage{tabularx}
\usepackage{multirow}
\usepackage{lscape}
\usepackage{subcaption}
\usepackage{epstopdf}

\usepackage{booktabs}

\usepackage{enumitem}

\usepackage{booktabs}
\usepackage{multirow}
\usepackage{array}

\usepackage{caption}
\usepackage{graphicx}
\usepackage{float}

\usepackage{subcaption}

\usepackage{float}
\usepackage{microtype}
\usepackage{lineno}
\usepackage{placeins}
\usepackage{xcolor}

\usepackage{algorithm}
\usepackage{algorithmic}

\newtheorem{remark}{Remark}

\usepackage{xcolor}
\usepackage{pifont}       
\usepackage{bbding}       
\newcommand{\cmark}{\ding{52}}
\newcommand{\xmark}{\ding{56}}

\usepackage{amsmath}
\renewcommand{\algorithmicrequire}{\textbf{Input:}}  
\renewcommand{\algorithmicensure}{\textbf{Output:}} 

\usepackage{threeparttable}
\usepackage{dcolumn}

\newcolumntype{d}[1]{D{.}{.}{#1}}
\newcommand{\tensor}[1]{\boldsymbol{\mathcal{#1}}}

\newtheorem{defn}{Definition}
\newtheorem{lemma}{Lemma}
\newcommand{\norm}[1]{\lVert#1\rVert}
\newcommand{\normlarge}[1]{\left\lVert#1\right\rVert}

\newcommand{\0}{\boldsymbol{\mathcal{O}}}
\newcommand{\1}{\mathbf{1}}

\newcommand{\C}{\boldsymbol{\mathcal{C}}}

\newcommand{\E}{\boldsymbol{\mathcal{E}}}
\newcommand{\G}{\boldsymbol{\mathcal{G}}}

\newcommand{\K}{\boldsymbol{\mathcal{K}}}

\newcommand{\M}{\boldsymbol{\mathcal{M}}}
\newcommand{\N}{\boldsymbol{\mathcal{N}}}

\newcommand{\W}{\boldsymbol{{\mathcal{W}}}}
\newcommand{\X}{\boldsymbol{\mathcal{X}}}
\newcommand{\Y}{\boldsymbol{\mathcal{Y}}}

\newcommand{\Am}{\boldsymbol{{A}}}

\newcommand{\Dm}{\boldsymbol{{D}}}

\newcommand{\Gm}{\textbf{{G}}}
\newcommand{\Qm}{\textbf{{Q}}}

\newcommand{\Sm}{\boldsymbol{S}}
\newcommand{\Xm}{\boldsymbol{X}}
\newcommand{\Ym}{\boldsymbol{Y}}

\newcommand{\Vm}{\boldsymbol{V}}
\newcommand{\Um}{\boldsymbol{U}}

\newcommand{\Zm}{\boldsymbol{Z}}

\newcommand{\Omegabm}{\boldsymbol{{\Omega}}}

\newcommand{\Pomega}{\boldsymbol{\mathcal{P}}_{\boldsymbol{{\Omega}}}}

\newcommand{\deltabm}{\boldsymbol{{\delta}}}

\ifCLASSINFOpdf
\else
\fi

\usepackage{algorithmic}
\usepackage{algorithm}
\renewcommand{\algorithmicrequire}{\textbf{Input:}}  
\renewcommand{\algorithmicensure}{\textbf{Output:}} 

\usepackage{xcolor}
\definecolor{darkblue}{rgb}{0.0,0.5,0.5}

\begin{document}
%

\title{Robust Tensor Completion via Gradient Tensor Nulclear $\ell_1$-$\ell_2$ Norm  for Traffic Data Recovery}

\author{\IEEEauthorblockN{Hao Shu, Jicheng~Li,~Tianyv~Lei,~Lijun~Sun}
}

\author{
Hao~Shu,~Jicheng~Li,~Tianyv~Lei,~Lijun~Sun
\thanks{The work was supported by National Natural Science Foundation of China (12171384). (\textit{Corresponding author: Jicheng Li})} 
\thanks{H. Shu, J. Li and T. Lei are with the School of Mathematics and Statistics, Xi'an Jiaotong University, Xi'an 710049, Shanxi, China (email: haoshu812@gmail.com,
 jcli@mail.xjtu.edu.cn,TianyvLei@stu.xjtu.edu.cn).}
\thanks{L.Sun is with the Department of Civil Engineering,
McGill University, Montreal, QC H3A 0C3, Canada. (email:  lijun.sun@mcgill.ca).}}

\maketitle
\begin{abstract}

In real-world scenarios, spatiotemporal traffic data frequently experiences dual degradation from missing values and noise caused by sensor malfunctions and communication failures. Therefore, effective data recovery methods are essential to ensure the reliability of downstream data-driven applications.
while classical tensor completion methods have been widely adopted, they are incapable of modeling noise, making them unsuitable for complex scenarios involving simultaneous data missingness and noise interference.
Existing \textit{Robust Tensor Completion} (RTC) approaches offer potential solutions by separately modeling the actual tensor data and noise. However, their effectiveness is often constrained by the over-relaxation of convex rank surrogates and  the suboptimal utilization of local consistency, leading to inadequate model accuracy.
To address these limitations, we first introduce the tensor  $\ell_1$-$\ell_2$ norm, a novel non-convex tensor rank surrogate that functions as an effective low-rank representation tool.
Leveraging an advanced feature fusion strategy, we further develop the gradient tensor
$\ell_1$-$\ell_2$ norm by incorporating the tensor  $\ell_1$-$\ell_2$   norm in the gradient domain. The rigorous mathematical analysis demonstrates that the gradient tensor $\ell_1$-$\ell_2$ norm not just serves as an effective low-rank regularizer but even possesses local consistency characterization capabilities.
By integrating the gradient tensor nuclear $\ell_1$-$\ell_2$ norm into the RTC framework, we propose the  \textit{Robust Tensor Completion via Gradient Tensor Nuclear  $\ell_1$-$\ell_2$ Norm} (RTC-GTNLN) model, which not only fully exploits both global low-rankness and local consistency without trade-off parameter, but also effectively handles the dual degradation challenges of missing data and noise in traffic data.
Extensive experiments conducted on multiple real-world traffic datasets demonstrate that the RTC-GTNLN model consistently outperforms existing state-of-the-art methods in complex recovery scenarios involving simultaneous missing values and noise.
The code is available at \url{https://github.com/HaoShu2000/RTC-GTNLN}.
\end{abstract}

\IEEEpeerreviewmaketitle

\section{Introduction}

\IEEEPARstart{T}he rapid advancement of communication and low-cost sensing technologies has facilitated a range of applications in Intelligent Transportation Systems (ITS), including traffic condition assessment \citep{yuan2012real}, real-time traffic management \citep{said2021spatiotemporal}, and long-term transportation planning \citep{zhang2021customized}. These applications fundamentally rely on the collection and analysis of vast amounts of traffic data, making the acquisition of complete and accurate traffic data crucial \citep{chen2024low}. However, data measurement errors such as signal interruptions or detector failures inevitably lead to missing values and noise, posing significant challenges to the effectiveness of ITS in utilizing and managing traffic data \citep{feng2022traffic}. Consequently, researchers are actively investigating methodologies that leverage prior knowledge of traffic patterns to address the prevalent challenge of traffic data recovery\citep{wang2018traffic,chen2019missing,chen2021bayesian,yu2025robust}.

\begin{figure*}[t]
\centering
\includegraphics[width=0.9\linewidth]{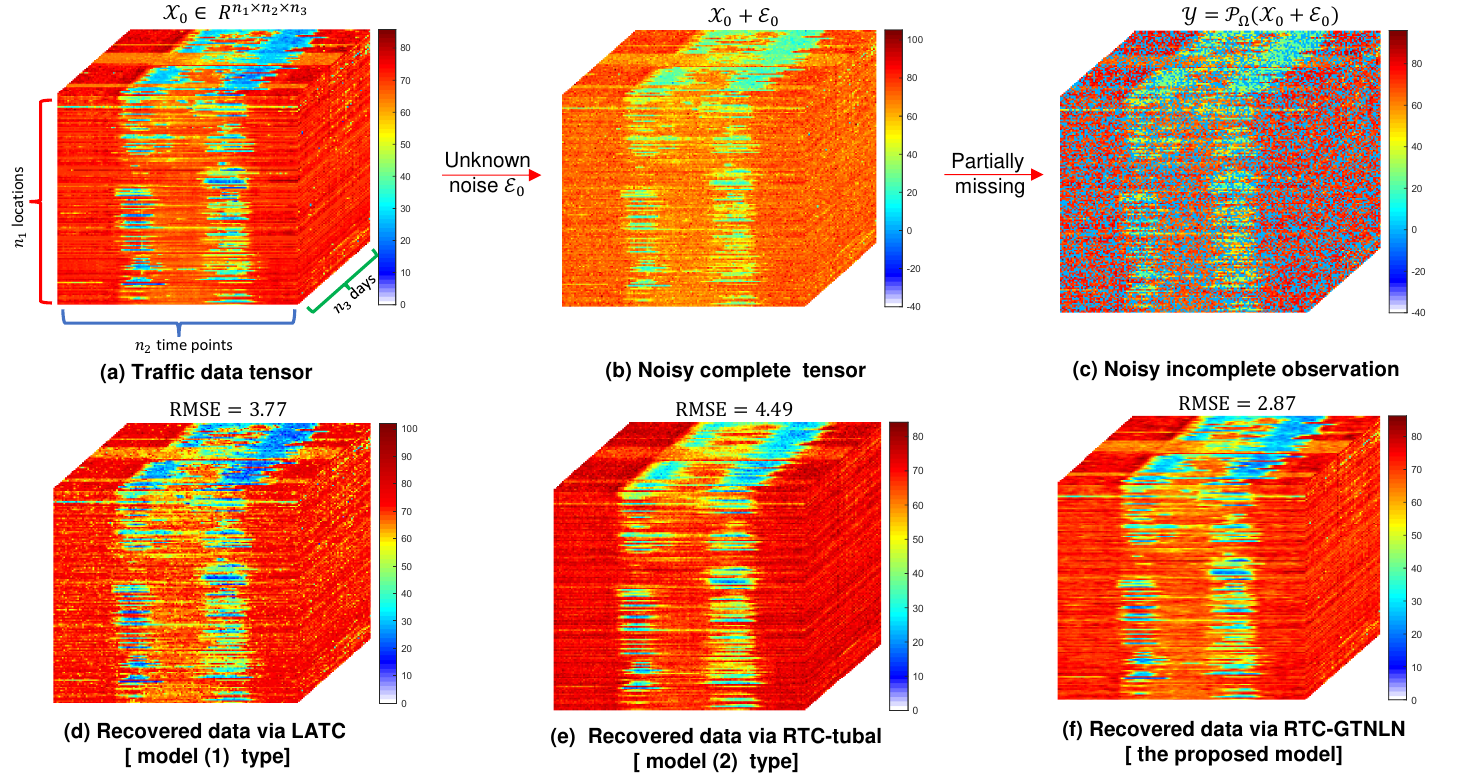}
\vspace{0.2cm}
\caption{Illustrations of the dual degradation of traffic data caused by missing values and noise, along with the recovery results from different models.  (a): Arrangement of spatiotemporal traffic tensor data; (b): noisy traffic tensor data affected by noise;  (c):  observed traffic tensor data impacted by both missing values and noise; (d): recovered tensor data using an efficient type (1) model; (e) recovered tensor data using a classical type (2) model; (f) recovered tensor data using the proposed model.}\label{ST}
\end{figure*}

When modeling multivariate spatiotemporal traffic data, a typical tensor representation (location × time × day, as shown in Fig.\ref{ST}(a)) offers a convenient method to capture complex relationships across multiple dimensions \citep{ran2016tensor}. Low-rank tensor completion methods, which encode the global structure of traffic data by minimizing the rank of the target tensor, have emerged as superior alternatives to traditional computational techniques  \citep{chen2020nonconvex,nie2022truncated,shu2024low}. Moreover, local consistency features, as important priors in spatiotemporal data, have been widely incorporated into low-rank tensor completion models, leading to the following general tensor model for processing traffic data:
\begin{equation}\label{type-1}
\begin{split}
&\min_{\tensor{X}}  ~  rank\left( \X \right)+ \theta \Psi (\X),\  \\
&\quad\text{s.t.} \ \Pomega(\X)= \Pomega(\X_0),
\end{split}
\end{equation}
where $\X_0$ represents the actual tensor traffic data,  $\Pomega(\cdot)$ is the projection operator associated with the observation set $\Omegabm$, the rank function $rank( \cdot )$ encodes the global low-rank prior information of the tensor $\X$, the term $\Psi (\cdot)$ denotes a regularizer that characterizes local consistency (e.g. total variation regularization\citep{chambolle2004algorithm}, fractional-order difference regularization \citep{zhang2012adaptive}, autoregressive-based temporal variation regularization \citep{chen2021low} and Laplacian kernel regularization \citep{chen2024laplacian}), and $\theta$  is the  trade-off parameter that balances global low-rankness and local consistency. Within this modeling framework, \cite{chen2021low} develop the state-of-the-art imputation method \textit{Low-Rank Autoregressive Tensor Completion}  (LATC) by utilizing the truncated nuclear norm as a rank surrogate and adding an autoregressive prior. However, the single imputation functionality limits the recovery capabilities of model (\ref{type-1}) in complex disruptive scenarios.

As depicted in Fig.\ref{ST}(b, c), traffic data often faces dual degradation—missing entries and noise—resulting in partially observed tensor being mixed with noise \citep{feng2022traffic,hu2023flexible}. When confronted with such scenarios, the recovery results of model (\ref{type-1}) still contain a significant amount of noise due to its lack of denoising capability, as shown in Fig.\ref{ST}(d). To address this issue, researchers have made significant strides in noise-aware decomposition  \citep{li2023robust,jiang2019robust,gao2020robust}.
By processing low-rank clean data and sparse noise separately,
the mathematical model addressing  complex disruptive  traffic data recovery problem can be formulated as a  \textit{Robust Tensor Completion} (RTC) model, i.e.,
\begin{equation}\label{type-2}
\begin{split}
&\ \min_{\X,\E}  ~  rank\left( \X \right)+\lambda \Upsilon(\E),\  \\
&\text{s.t.} \ \Pomega(\X+\E)= \Pomega(\Y), 
\end{split}
\end{equation}
Where $\Y=\Pomega(\X_0+\E_0)$ denotes the observed traffic data tensor, the noise tensor $\E_0$ reflects the noise signal,  the sparsity measure $\Upsilon(\cdot)$ is used to separate the noise tensor $\E$ from the actual traffic data,  and  $\lambda$ is an empirical value determined solely by tensor dimensions that requires no tuning \citep{song2020robust,hu2023flexible}.
However, a series of methods in model category (\ref{type-2}) solely characterize the global low-rankness of the data tensor \citep{jiang2019robust,song2020robust,hu2023flexible,hu2024vehicle}. For instance, the classic robust tensor completion  method RTC-tubal \citep{jiang2019robust} is constructed based on the tensor tubal rank within the tensor singular value decomposition framework. As illustrated in Fig.\ref{ST}(e), employing an over-relaxed convex surrogate for tensor tubal rank and disregarding the local consistency of traffic data, the recovered tensor from  simple RTC model demonstrates insufficient precision in its detailed information.
The potential directions for methodological improvement are multifaceted. On one hand, replacing the convex tensor nuclear norm with more effective nonconvex tensor nuclear norm alternatives could better approximate tensor rank characterization\citep{hu2023flexible}. On the other hand, 
adding local consistency constraints to the data recovery framework (\ref{type-2}) shows practical promise \citep{feng2022traffic}.
However, simply incorporating the local consistency regularizer into the RTC model, i.e., 
 \begin{equation}\label{type-3}
 \begin{split}
 \min_{\X,\E}  ~&  rank\left( \X \right)+ \theta \Psi (\X)+\lambda \Upsilon(\E),\  \\
& \text{s.t.} \ \Pomega(\X+\E)= \Pomega(\Y).
 \end{split}
\end{equation}
decouples the modeling of local consistency and global low-rankness. This not only fails to fully exploit their synergy for performance gains but also introduces the complex hyperparameter $\theta$ to strike the balance between global low-rankness and local consistency. (See the examination of model (\ref{separate}) in the ablation study for details).
Therefore, the parameter-free precise integration of local consistency and global low-rankness based on non-convex nuclear norm remains an unresolved and urgent problem in robust tensor completion frameworks.


To address this problem, we propose a method called \textit{Robust Tensor Completion via Gradient Tensor Nuclear  $\ell_1$-$\ell_2$  Norm} (RTC-GTNLN), which achieves high-precision denoising and imputation simultaneously (see Fig.\ref{ST}(f) for verification). The method is designed to recover complex corrupted traffic data by developing an effective non-convex tensor rank surrogate  and integrating local consistency into the robust low-rank tensor completion model in an inseparable manner. Concretely,  we first design the
\textit {Tensor Nuclear $\ell_1$-$\ell_2$ Norm}(TNLN) as a novel and effective non-convex alternative for tensor Tucker rank. In addition, we jointly encode global low-rankness and local consistency by representing the low-rankness of the gradient tensor. We mathematically prove that  the \textit{Gradient Tensor
Nuclear $\ell_1$-$\ell_2$  Norm}(GTNLN)  not only serves as a low-rank regularizer, but also provides smoothness characterization capability  similar to the total variation. Finally, relying on the framework of robust tensor completion, we decompose the observed tensor into a low-rank actual component and a sparse noise component, achieving simultaneous denoising and missing value imputation.

Notably, our model is highly user-friendly as it avoids separating global low-rank regularization and local consistency regularization as in the current work \citep{feng2022traffic,chen2021scalable,chen2021low,zeng2025flexible,zhang2025symmetric}, thereby eliminating trade-off parameter tuning. 
In the implementation of the proposed RTC-GTNLN, we design an efficient algorithm based on the \textit{Alternating Direction Method of Multipliers} (ADMM) framework, where all variables are updated using closed-form solutions.
While the RTC-GTNLN model integrates more priors compared to RTC models with  only low-rank prior, our analysis shows that the proposed algorithm maintains efficient time complexity through a fast sub-iteration solver.
We further validate the method through extensive numerical experiments on real-world traffic datasets, highlighting its effectiveness in degraded data recovery. Overall, the main contributions of this work are summarized as follows:
\begin{itemize}
\item Inspired by the effectiveness of the $\ell_1$-$\ell_2$ norm in compressed sensing and matrix rank approximation, we propose a novel and efficient non-convex tensor rank surrogate, termed TNLN. Building upon this, we further introduce a regularizer called GTNLN, which simultaneously encodes both global low-rankness and local consistency through advanced feature fusion techniques.

\item A non-convex high-precision traffic data recovery model RTC-GTNLN is proposed, which not only fully leverages global low-rankness and local consistency in a parameter-free manner, but also achieves simultaneous missing data interpolation and noise removal.


\item The proposed RTC-GTNLN model is efficiently solved within the ADMM framework and rigorously evaluated on real-world traffic datasets. Experimental results demonstrate that our method achieves superior recovery accuracy in challenging scenarios involving both noise and missing data, outperforming state-of-the-art algorithms in related works.

\end{itemize}
The structure of this paper is outlined as follows.  Section~\ref{sec:rel} provides an overview of relevant studies. Next, Section~\ref{sec:notations and preliminaries} establishes the necessary notations and foundational concepts. Section~\ref{sec:Method and Model} then introduces the proposed RTC-GTNLN method, while Section~\ref{sec:Algorithm} details the associated optimization algorithm. Section~\ref{sec:experiments} presents comprehensive experiments conducted on multiple traffic datasets, along with comparisons against several baseline models. The paper concludes with Section~\ref{sec:conclusion}, which summarizes the findings and discusses their implications.

\section{Related Works}\label{sec:rel}
This section reviews related work in three aspects: tensor rank and its surrogates for global low-rankness, low-rank tensor completion with local consistency, and traffic data recovery from noisy partial observations. 
\subsection{Tensor Rank and Its Surrogates for Global Low-rankness}
Global low-rankness is a prominent characteristic of spatiotemporal traffic data \citep{li2024convolutional,chen2019bayesian,lyu2024tucker,baggag2019learning,ling2021t,yang2023transforms,chen2024nt}, and its representation often relies on the exploration of tensor ranks and their surrogates. Unlike matrix rank, the definition of tensor rank is not unique and varies with different tensor decomposition methods. Common tensor ranks include the CP rank \citep{hitchcock1927expression}, Tucker rank \citep{kolda2009tensor}, tubal rank \cite{lu2019low}, train rank \citep{oseledets2011tensor} and ring rank \citep{zhao2016tensor}.
Directly minimizing these tensor ranks is challenging—for instance, accurately estimating the CP rank is mathematically proven to be an NP-hard problem \citep{hillar2013most}. 

To systematically exploit global low-rankness in tensor modeling, researchers adopt a series of convex relaxations as substitutes for tensor rank. For example, \cite{liu2012tensor} employ  the 
sum of nuclear norms as a proxy for the Tucker rank, while \cite{zhang2016exact} propose the tubal nuclear norm as a surrogate for the tubal rank. However, the over-relaxation introduced by these convex nuclear norms limits their ability to closely approximate tensor ranks, thereby affecting the accuracy of low-rank models. This sparks a series of studies on non-convex rank surrogates \citep{chen2020nonconvex,nie2022truncated,gao2020robust,hu2023flexible,hu2012fast}.
 These non-convex tensor rank surrogates are obtained by applying certain non-convex penalty functions to the singular values, such as $\ell_p$ norm \citep{gao2020robust}, truncated function \citep{hu2012fast},  capped $\ell_1$ norm \citep{zhang2010analysis}, variant logarithmic function \citep{friedman2012fast}, Laplace function \citep{trzasko2008highly}, 
  ETP \citep{gao2011feasible}, MCP \citep{zhang2010nearly}, SCAD \citep{fan2001variable}. For instance, \cite{chen2020nonconvex}  introduce the truncated nuclear norm as a non-convex surrogate for Tucker rank with demonstrated recovery efficacy, while \cite{nie2022truncated} subsequently extend this nuclear norm  by proposing the truncated Schatten $p$-norm for enhanced tensor completion performance.
 In summary, the tensor nuclear norm, particularly its non-convex variants as effective rank surrogates, has been widely applied in tensor rank approximation and has demonstrated significant utility in tensor data modeling.

\subsection{Low-Rank Tensor Completion with Local Consistency}
In low-rank tensor completion models for traffic data recovery, local consistency is a widely validated and utilized prior knowledge. Common tools for characterizing local consistency include total variation \citep{chambolle2004algorithm}, autoregressive-based temporal variation \citep{chen2021low}, Laplacian kernel regularization \citep{chen2024laplacian}, generalized temporal consistency regularization \citep{nie2023correlating}, fractional-order difference regularization \citep{zhang2012adaptive}, and long-short-term neural networks \citep{yang2021real}.
However, existing methods  usually rely on model (\ref{type-1}) to separately characterize global low-rankness and local consistency \citep{chen2021scalable,chen2021low,zeng2025flexible,nie2023correlating,lyu2024tucker,ling2021t,chen2024laplacian,lei2024generalized}. The performance of such models heavily relies on the selection of trade-off parameter. In many cases, determining the optimal trade-off parameter remains difficult and often varies with different datasets, which significantly restricts the real-world implementation of these approaches.


Recent advances in tensor modeling see the emergence of regularization techniques that simultaneously encode global low-rankness and local consistency \citep{peng2022exact,wang2023guaranteed,shu2024low}. Specifically, \cite{peng2022exact} and \cite{wang2023guaranteed} introduce nuclear norms applied to first-order difference tensors instead of raw data matrices, demonstrating theoretically that their proposed correlated total variation regularizers inherently combine smoothness with low-rank properties. However, the over-relaxation of rank constraints caused by convex nuclear norms fundamentally limits reconstruction precision. 
Building upon these foundations, \cite{shu2024low} develop the LRTC-3DST framework through non-convex nuclear norm regularization across multiple localized feature tensors for spatiotemporal traffic data imputation. 
While this architecture demonstrates state-of-the-art recovery accuracy in missing value estimation tasks, its reliance on geographic location information and significant computational overhead hinder practical deployment in traffic management systems. 
\begin{table*}[!t]
	\centering
	\caption{Summary of some existing tensor completion models for traffic data recovery.}\label{tab_low}
\resizebox{\textwidth}{!}{
	\begin{tabular}{c||c|c|c|c|c}
  Model   & Global low-rankness & non-convex rank surrogate &  Local consistency
     &No trade-off parameter $\theta$? & Noise robustness \\ \hline
HaLRTC \citep{ran2016tensor}  & the sum of
nuclear norms  & \xmark  & \xmark
 & \cmark          & \xmark    \\ \hline
  LRTC-TNN \citep{chen2020nonconvex} &
 truncated  nuclear norm  & \cmark & \xmark
 & \cmark                 & \xmark             \\ \hline
 LRTC-TSpN  \citep{nie2022truncated} &
truncated Schatten $p$-norm  & \cmark & \xmark
 & \cmark                 & \xmark             \\ \hline
 LSTC-tubal \citep{chen2021scalable}  & tubal nuclear norm   & \xmark   & \cmark & \xmark                  & \xmark   \\ \hline
LETC \citep{nie2023correlating}   &  graph tensor nuclear norm    & \xmark & \cmark & \xmark                  & \xmark   \\ \hline

   LATC \citep{chen2021low} &
 truncated  nuclear norm  & \cmark & \cmark
 & \xmark                 & \xmark             \\ \hline
  GCLSNL  \citep{zeng2025flexible} & mode-$k$ tubal nuclear norm & \xmark & \cmark
 & \xmark                 & \xmark             \\ \hline

    LRTC-3DST \citep{shu2024low}  & truncated nuclear norm  & \cmark & \cmark
 & \cmark                    & \xmark         \\ \hline
    LRTC-TCTV  \citep{wang2021generalized} & tubal nuclear norm & \xmark  & \cmark
 & \cmark                    & \xmark         \\ \hline
 RSCPN  \citep{hu2023flexible}  & tensor Schatten capped $p$-norm & \cmark  & \xmark & \cmark                  & \cmark   \\ \hline
 LFTC \citep{hu2024vehicle} & truncated nuclear norm  & \cmark  & \xmark & \cmark                  & \cmark   \\ \hline
 ST-TRPCA \citep{feng2022traffic} & tubal nuclear norm & \xmark   & \cmark & \xmark                  & \cmark   \\ \hline
 RTC-SPN  \citep{gao2020robust}  & Schatten $p$-norm  & \cmark  & \xmark & \cmark                  & \cmark   \\ \hline
 RTC-tubal   \citep{jiang2019robust} & tubal nuclear norm & \xmark   & \xmark & \cmark                  & \cmark   \\ \hline
 RTC-TTSVD  \citep{song2020robust}& transformed tubal nuclear norm & \xmark   & \xmark & \cmark                  & \cmark   \\ \hline
 RTC-GTNLN (This work) &  tensor
nulclear $\ell_1$-$\ell_2$ norm  & \cmark  & \cmark
 & \cmark                    & \cmark         \\ \hline    
	\end{tabular}}  
\end{table*}

\subsection{Traffic data recovery from noisy partial observations}
 Traffic data collected from a variety of heterogeneous sources often suffers from data incompleteness and measurement errors caused by factors such as sensor failures, software malfunctions, and network disruptions \citep{feng2022traffic,hu2023flexible,hu2024vehicle}. This makes missing and noisy traffic data a common occurrence. While tensor completion models have been widely and systematically studied for traffic data imputation, the current range of tensor completion methods is not well-suited to handle the dual degradation of missing values and noise \citep{chen2021bayesian,chen2020nonconvex,chen2021scalable,li2024convolutional,chen2019bayesian}.
 
At present, robust tensor completion methods for recovering incomplete and noisy traffic data remain less advanced and comprehensive compared to standard tensor completion models, particularly in terms of mining and fusing data features \citep{hu2023flexible,jiang2019robust,gao2020robust,hu2024vehicle,song2020robust}.
For instance, \cite{hu2023flexible,hu2024vehicle} incorporate the truncated nuclear norm and the Schatten capped $p$-norm into the RTC framework but overlook the local consistency of spatiotemporal traffic tensors.
\cite{feng2022traffic} introduce spatiotemporal constraint terms into the robust tensor decomposition framework, resulting in their proposed ST-TRPCA model requiring predefined tensor ranks and prior-balancing hyperparameters.

Generally, some existing traffic data recovery methods based on tensor completion are summarized in Table \ref{tab_low}. 
Our proposed RTC-GTNLN model not only achieves the joint encoding of global low-rankness and local consistency without parameter trade-off through the non-convex regularizer GTNLN, but also effectively solves the dual degradation problems of missing data and noise in traffic data.

\section{Notations and Preliminaries}\label{sec:notations and preliminaries}

Throughout this paper, we use lowercase, boldface lowercase, boldface capital and boldface Euler script letters to denote scalars, vectors, matrices and tensors, respectively, e.g., $x \in \mathbb{R}$, $\boldsymbol{x}\in \mathbb{R}^{n}$, $\boldsymbol{X}\in \mathbb{R}^{n_1 \times n_2}$, and $\tensor{X} \in \mathbb{R}^{n_1 \times n_2 \times  n_3}$. The $(i_1,i_2,i_3)$-th entry of a third-order tensor is denoted by $\tensor{X}_{i_1i_2i_3}$. 
The element-wise $\ell_0$ norm,  $\ell_1$ norm, and $\ell_2$ norm (i.e., Frobenius norm) of a third-order real tensor are defined as $\|\tensor{X}\|_0= \sharp\left\{(i_1,i_2,i_3): \tensor{X}_{i_1i_2i_3}\neq 0\right\}$( $\sharp$ represents the number of elements in the set),
 $\|\tensor{X}\|_1 = \sum_{i_1i_2i_3}|\tensor{X}_{i_1i_2i_3}|$, 
$\|\tensor{X}\|_F = \sqrt{\sum_{i_1i_2i_3}   \tensor{X}_{i_1i_2i_3 }^2}$, respectively.  The element-wise $\ell_0$ norm,  $\ell_1$ norm, and $\ell_2$ norm of vectors and matrices are similar to those of tensors.
The $i$-th mode ($i=1,2,3$) unfolding matrix $\Xm_i$ of a third-order tensor $\boldsymbol{\mathcal{X}}$ is obtained through the unfold operator
$\operatorname{unfold}_{i}(\X)=\Xm_i$.
,and the fold operator $\operatorname{fold}_{i}(\cdot)$ converts a matrix back to a third-order tensor along the $i$-th mode, i.e., $\operatorname{fold}_{i}(\Xm_i)=\boldsymbol{\mathcal{X}}$. 
The $i$-th mode product of a third-order tensor $\X$ with a matrix $\Dm $ is denoted by $\X \times_i \Dm$, and $\Y=\X \times_i \Dm$ is equivalent to $\Ym_i=\Dm \Xm_i$ \cite{kolda2009tensor}. 
For a rank-$r$ matrix $\mathbf{X}$, define its singular value vector as
$\boldsymbol{\sigma}(\mathbf{X}) \triangleq \left(\sigma_1(\mathbf{X}), \sigma_2(\mathbf{X}), \ldots, \sigma_r(\mathbf{X})\right)^{\mathrm{T}}$,
where $\sigma_i(\mathbf{X})$ denotes the $i$-th singular value ordered by magnitude. The matrix nuclear norm can be expressed as the $\ell_1$ norm of its singular value vector, i.e.,
$\norm{\Xm}_* = \norm{\boldsymbol{\sigma}(\Xm)}_1 = \sum_{i=1}^r \sigma_i(\Xm)$.

\section{Proposed TNLN-TGD Model}

\label{sec:Method and Model}
This section presents the RTC-GTNLN model through three key components: 1) Low-rank prior representation: Building upon the low-rank characteristics of traffic data, we design a tensor nulclear $\ell_1$-$\ell_2$ norm as a non-convex surrogate for tensor rank estimation. 2) Gradient domain constraint: In the temporal gradient domain, we further introduce a gradient nulclear $\ell_1$-$\ell_2$ norm to jointly encode both global low-rankness and local consistency. 3) Noise-aware decomposition: The observed tensor is decomposed into a low-rank ground-truth component and a sparse noise component, achieving simultaneous denoising and missing-value imputation.


\subsection{Low-rank Prior Representation for Global Consistency }






For a tensorial traffic data $\X\in\mathbb{R}^{n_1\times n_2\times n_3}$ with $n_1$ locations, $n_2$ time points and $n_3$ days,  its Tucker rank is defined as the vector of ranks from mode-unfolded matrices.
The weighted Tucker rank, extensively employed in tensor modeling, is formulated as:
\begin{equation}\label{r_rankt}
\operatorname{rank}_t{(\X)}\triangleq\sum_{i=1}^{3} \alpha_i \operatorname{rank}{(\Xm_i)} =\sum_{i=1}^{3}\alpha_i \norm{\boldsymbol{\sigma}(\Xm_i)}_0
\end{equation}
where $\Xm_i$ denotes the mode-$i$ unfolding matrix of tensor $\X$,  the weighting factors $\alpha_i(i=1,2,3)$ satisfy the normalization condition $\sum_{i=1}^3 \alpha_i=1$,
 $\boldsymbol{\sigma}(\Xm_i)$  represents the singular value vector of $\Xm_i$, and $\norm{\cdot}_0$ denotes the $\ell_0$ norm, which counts the number of nonzero elements in a vector.
It can be observed that the low-rank structure of a tensor corresponds to the sparsity of its singular values. In other words, when the number of zero singular values is very large, or when the energy of the singular value vector is concentrated in the first few singular values, the tensor exhibits strong low-rankness.
By introducing a convex surrogate to approximate the weighted Tucker rank, \cite{liu2012tensor} obtain the following tensor nuclear norm:
\begin{equation}\label{r_TNN}
\norm{\X}_{\circledast}\triangleq\sum_{i=1}^{3} \alpha_i  \|\Xm_i\|_{*}=\sum_{i=1}^{3}\alpha_i \norm{\boldsymbol{\sigma}(\Xm_i)}_1
\end{equation}
where $\norm{\X}_{*}$ is the matrix nuclear norm and $\norm{\cdot}_1$ represents the $\ell_1$ norm. Compared to the weighted Tucker rank, the tensor nuclear norm employs the convex  $\ell_1$ norm as a proxy instead of the $\ell_0$ norm, significantly improving solvability. However, the $\ell_1$ norm's over-relaxation relative to the  $\ell_0$ norm makes it an insufficient approximation of the Tucker rank, compromising the precision of tensor modeling \citep{nie2022truncated}.

\begin{figure}[t]
\renewcommand{\arraystretch}{0.5}
\centering
\begin{tabular}{ccccccc}
\centering
\includegraphics[width=44mm, height = 40mm]{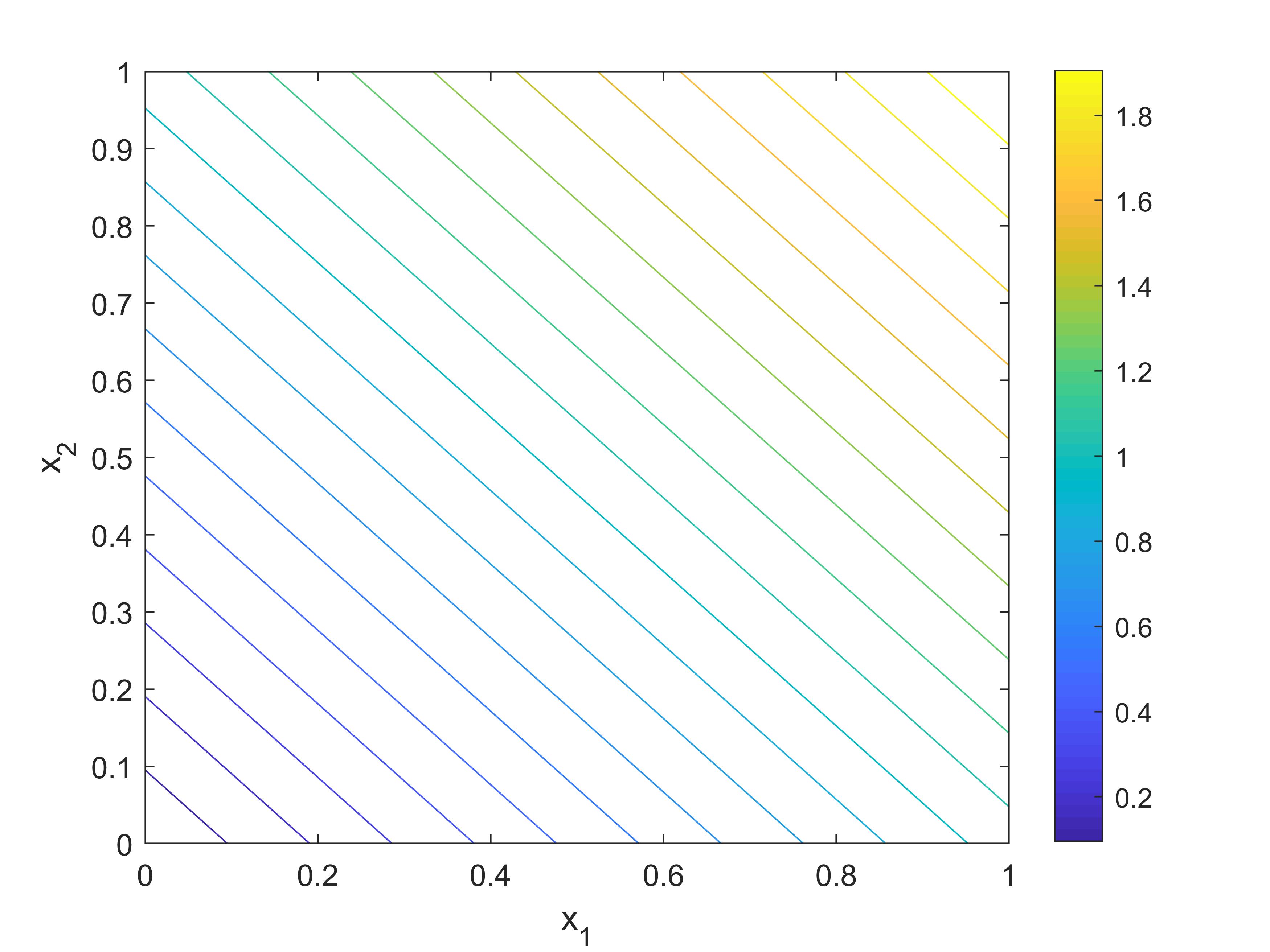}&  
\includegraphics[width=44mm, height = 40mm]{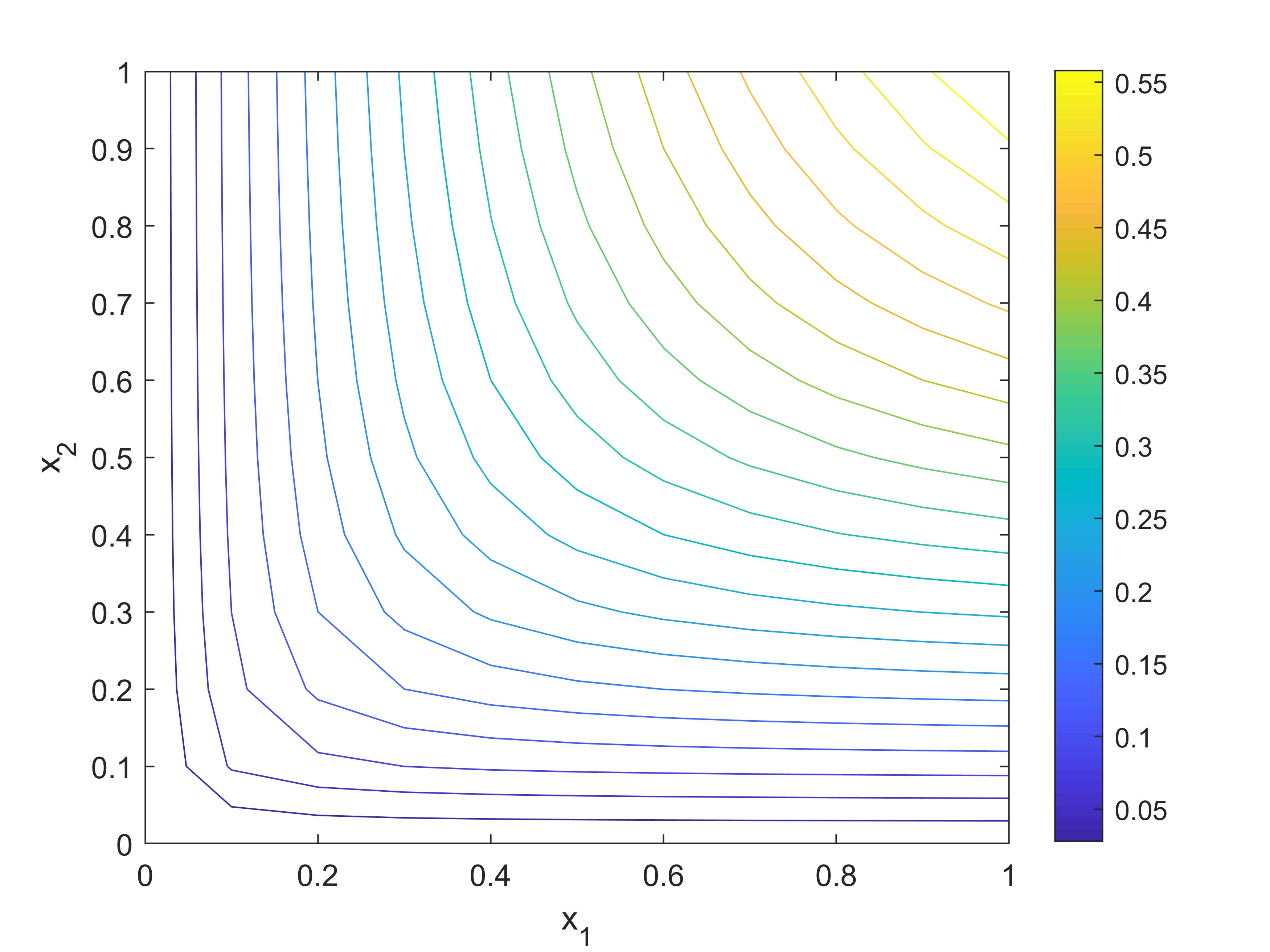}\\
\scriptsize \textbf{$\ell_1$ norm}& \scriptsize \textbf{$\ell_1$-$\ell_2$ norm}  \\
\end{tabular}
\caption{ Contour plot of the $\ell_1$ norm  and $\ell_1$-$\ell_2$  norm  of vector $\mathbf{x}=(x_1,x_2) $.}\label{fig.normL12}
\end{figure}

In the field of compressed sensing, the $\ell_1$-$\ell_2$ norm serves as a reliable non-convex surrogate for the $\ell_0$ norm \citep{yin2015minimization,lou2018fast}. As illustrated in Fig.\ref{fig.normL12}, the contour lines of the $\ell_1$-$\ell_2$  norm exhibit a closer proximity to the coordinate axes compared to those of the $\ell_1$ norm. This geometric characteristic enables the minimization of the $\ell_1$-$\ell_2$  norm to yield solutions with enhanced sparsity. More recently, the $\ell_1$-$\ell_2$ norm has been successfully applied to enhance the matrix nuclear norm, demonstrating promising results in matrix completion tasks \citep{li2024novel}. This advancement further validates its superior performance over the $\ell_1$ norm in multivariate data modeling. Motivated by this, we propose the tensor nuclear $\ell_1$-$\ell_2$ norm.  

\begin{defn}
\label{definition1}
(Tensor Nulclear  $\ell_1$-$\ell_2$  Norm (TNLN)). For a third-order tensor, its tensor nulclear  $\ell_1$-$\ell_2$  norm is defined as
\begin{equation}
    \norm{\X}_{\circledast,\ell}\triangleq\sum_{i=1}^{3}\alpha_i (\norm{\boldsymbol{\sigma}(\Xm_i)}_1-\norm{\boldsymbol{\sigma}(\Xm_i)}_2)
\end{equation}
where $\boldsymbol{\sigma}(\Xm_i)$ denotes the singular value vector of the unfolding matrix $\Xm_i$, $\norm{\cdot}_1$ and $\norm{\cdot}_2$ represent the $\ell_1$ norm and  $\ell_2$ norm respectively, and the weight coefficients $\alpha_i(i=1,2,3)$ satisfy the constraint $\sum_{i=1}^3 \alpha_i=1$.
\end{defn}

Currently popular rank non-convex surrogates often require certain hyperparameters for control, such as the truncation coefficient of truncated nuclear norm \cite{chen2020nonconvex},  the norm exponent of  Schatten $p$-norm \cite{gao2020robust}, and the threshold parameter of  Schatten capped $p$-norm \cite{hu2023flexible}. The TNLN we propose serves as a reliable non-convex surrogate that does not require any hyperparameters, making it more user-friendly in practice.

\subsection{Gradient Domain Constraint for Local Consistency }
\begin{figure}[t]
\centering
\includegraphics[width=1\linewidth]{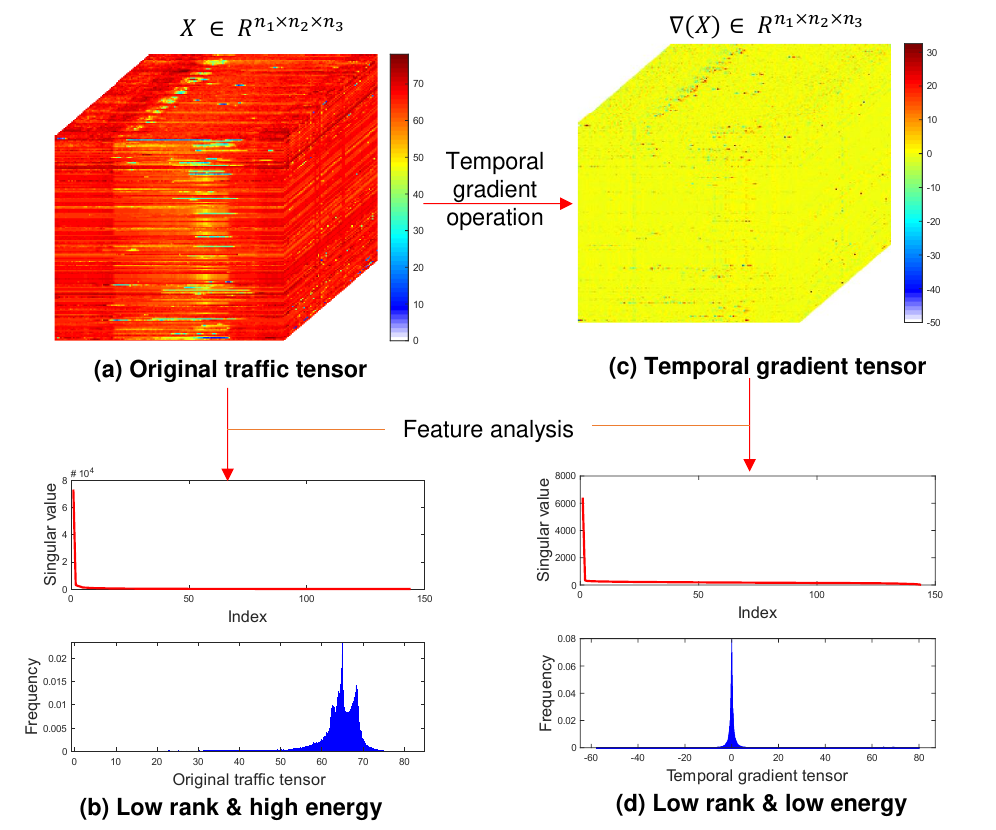}
\caption{Illustrations of characteristics in original traffic data and its  gradient tensor. (a): Original traffic tensor in the real world. (b): 
Singular value curves of the original traffic tensor (upper). The rapid decline in singular values indicates that the traffic tensor has a low-rank structure.
Frequency histograms of all elements in the tensor (below). The distribution of elements across higher values suggests that the original tensor is characterized by high energy.
(c): temporal gradient tensor via gradient operation;  (d): Singular values curves of temporal gradient tensor (upper) and frequency histograms of all their elements (below), the rapid decrease of singular values indicates that the gradient tensor is of low rank, and the elements concentrated near 0 indicate that the gradient tensor is of low energy.
}\label{fig:gradient tensor}
\end{figure}

Considering traffic data  $\X\in\mathbb{R}^{n_1\times n_2\times n_3}$  with global consistency and local consistency priors, its global consistency can be well characterized via minimizing the rank surrogates, like the  tensor nulclear  $\ell_1$-$\ell_2$ norm.  Its local consistency, which is often captured by the temporal anisotropic \textit{Total Variation} (TV) \citep{wang2018traffic,ling2021t,chen2021scalable}, is defined as:
\begin{equation}
    \norm{\X}_{TV}\triangleq \norm{ \Dm\Xm_2}_F= \norm{\X \times_2\Dm}_F = \norm{\nabla(\X)}_F
\end{equation} 
where $\nabla(\X)=\X \times_2\Dm$ is the temporal gradient tensor, $\Dm \in\mathbb{R}^{n_2\times n_2}$ is a row circulant matrix of $(-1,1,0,\cdots,0)$.
As shown in the Fig.\ref{fig:gradient tensor}, the temporal gradient tensor exhibits both  low energy (most values are close to 0) and low rank (fast decaying singular values)   properties. Unlike the traditional locally consistent regular TV, which captures the low-energy property  of  temporal gradient tensor $\nabla(\X)$  through the Frobenius norm constraint, we impose TNLN on $\nabla(\X)$  to encode its low-rankness, thereby proposing  the gradient tensor nulclear  $\ell_1$-$\ell_2$  norm.


\begin{defn}
\label{definition2}
(Gradient Tensor Nulclear  $\ell_1$-$\ell_2$  Norm (GTNLN)). For a third-order tensor $\X$ , its gradient  tensor nulclear  $\ell_1$-$\ell_2$  norm is defined  as
\begin{equation}
    \norm{\nabla(\X)}_{\circledast,\ell}\triangleq\sum_{i=1}^{3}\alpha_i (\norm{\boldsymbol{\sigma}(\Gm_i)}_1-\norm{\boldsymbol{\sigma}(\Gm_i)}_2)
\end{equation}
where $\Gm_i$ is the mode-$i$ unfolding matrix of the gradient  tensor $\G=\nabla(\X)$, $\boldsymbol{\sigma}(\Gm_i)$ is the singular value vector of  $\Gm_i$, and $\alpha_i(i=1,2,3)$ are weight coefficients satisfying $\sum_{i=1}^3 \alpha_i=1$. 
\end{defn}
Interestingly, by characterizing the gradient tensor $\nabla \X$, the proposed GTNLN demonstrates dual capability in encoding both global low-rankness and local consistency of original traffic tensor $\X$.
\begin{remark}
For a tensorial traffic data $\X\in\mathbb{R}^{n_1\times n_2\times n_3}$ with $n_1$ locations, $n_2$ time points and $n_3$ days, then $\norm{\nabla(\X)}_{\circledast,\ell}$ can characterize its global low-rankness  and local consistency simultaneously. The reasons are as follows: 1) On the one hand, since the linear transform $\Dm$, which connects the temporal gradient tensor $\nabla \X$ with the original tensor $\X$, is approximately full-rank, the rank of $\nabla \X$  remains consistent with that of $\X$. Consequently, the low-rank constraint applied to $\nabla \X$ ensures the global low-rankness of $\X$.
2) On the other hand, the tensor nuclear $\ell_1$-$\ell_2$  norm possesses an intrinsic monotonic property, which enables the regulation of numerical values within through its minimization. This mechanism effectively captures the numerical decay patterns in the temporal gradient tensor $\nabla \X$, consequently representing the local consistency in the traffic data.
\end{remark}

To rigorously demonstrate that the proposed GTNLN, as a nuclear norm constraint, can function similarly to TV, we present the following lemma:
\begin{lemma} \label{GTNLN-TV}
For traffic tensor $\X\in\mathbb{R}^{n_1\times n_2\times n_3}$ with $n_1$ locations, $n_2$ time points and $n_3$ days, it can be verified that
\[(\sqrt{1+\frac{1}{\eta(\G)}}-1) \norm{\X}_{TV} \leq \norm{\nabla(\X)}_{\circledast,\ell} \leq(\sqrt{r}-1)\norm{\X}_{TV}, \]
where $\G=\nabla \X$,  $\Gm_i(i=1,2,3)$ is  unfolding matrix of $\G$, $r=\operatorname{max} (r_1,r_2,r_3)$, $r_1=rank(\Gm_1)$, $r_2=rank(\Gm_2)$, $r_3=rank(\Gm_3)$,
$\eta_0(\Gm_i)=\operatorname{max}(\frac{\sigma_1(\Gm_i)}{\sigma_2(\Gm_i)},\frac{\sigma_2(\Gm_i)}{\sigma_3(\Gm_i)},\cdots,\frac{\sigma_{r_0-1}(\Gm_i)}{\sigma_{r_0}(\Gm_i)})$, and $\eta(\G)=\operatorname{max}(\eta_0(\Gm_1),\eta_0(\Gm_2),\eta_0(\Gm_3))$.


\end{lemma}

The proof of Lemma \ref{GTNLN-TV} can be found in the Appendix. Lemma \ref{GTNLN-TV} reveals that GTNLN and TV are compatible in terms of norms. Minimizing the GTNLN regularization will result in a smaller TV, thereby constraining the local consistency. In summary, The proposed GTNLN serves as a unified regularizer that simultaneously characterizes both global low-rankness and local consistency, thereby eliminating the trade-off parameter between these two regularization terms.

\subsection{Noise-aware Decomposition for Completion Robustness}

In the collection and transmission of traffic data, not only do missing values exist, but various types of noise are frequently encountered, such as Laplacian noise caused by abnormal events \citep{khatua2018sparse}, Gaussian noise resulting from sensor measurement errors \citep{wu2024traffic}, and hybrid composite noise arising from the combined influence of both types of noise \citep{peng2024stable}. 

\begin{figure}[!htbp]
\centering
\includegraphics[width=1\linewidth]{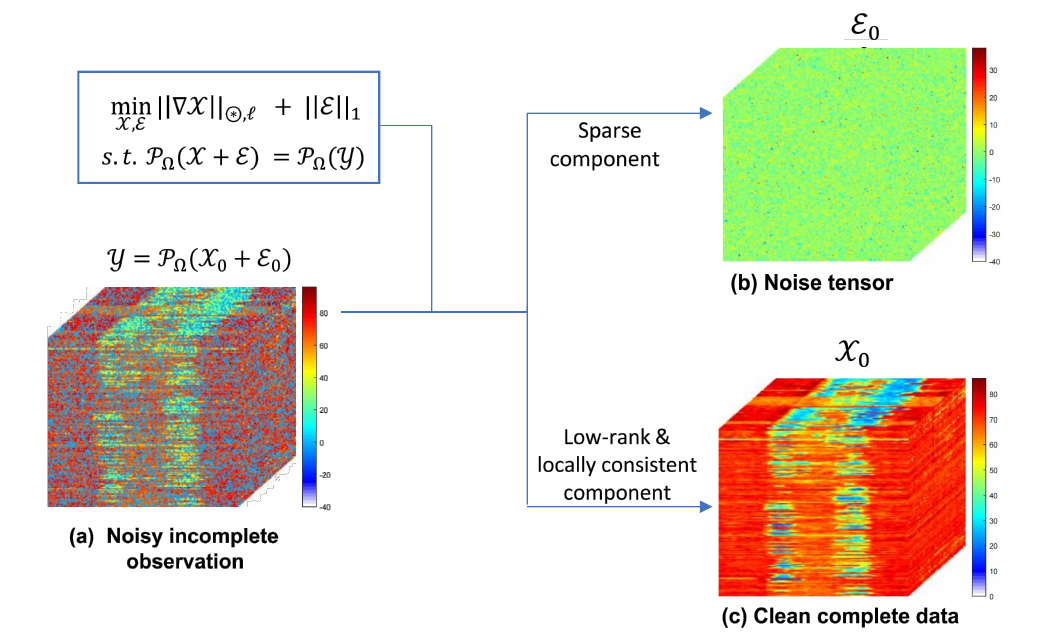}
\caption{RTC-GTNLN's dual capability: A PEMS07 traffic Data case study in imputation and denoising. (a) Noisy incomplete observation; (b) sparse noise tensor; (c) recovered clean complete data via RTC-GTNLN.
}\label{fig:noise_decom}
\end{figure}

Given the dual corruption of noise and missing values, the observed traffic tensor can be represented as $\Y=\Pomega(\X_0+\E_0)$, where $\X_0$ denotes the actual  traffic data, $\mathcal{E}_0$ corresponds to the sparse noise component, and $\mathcal{P}_{\Omega}(\cdot)$ represents the restriction operator defined over the sampling index set $\Omega$.  
To address this dual degradation problem, we introduce a noise-aware decomposition strategy that models the noise tensor and the actual traffic tensor separately  based on their distinct characteristics, thereby establishing the following robust tensor recovery framework for noisy incomplete traffic data recovery.
\begin{equation}\label{RTC}
\begin{aligned}
\min_{\X,\E}  ~  \mathcal{R} \left( \X \right)+\lambda \Upsilon(\E),\ \\
\text{s.t.} \ \Pomega(\X+\E)= \Pomega(\Y),
    \end{aligned}
\end{equation}
where  $\mathcal{R}(\cdot)$  denotes the regularizer meticulously formulated to simultaneously encapsulate the global low-rank structure and local consistency inherent in the original traffic tensor,  while 
$\Upsilon(\cdot)$ rigorously quantifies the sparse characteristics of the noise component. By employing GTNLN as a feature regularizer for the traffic tensor 
we  propose the \textit{Robust Tensor Completion via Gradient Tensor Nuclear  $\ell_1$-$\ell_2$  Norm} (RTC-GTNLN) model:
\begin{equation}\label{RTC-GTNLN}
\begin{aligned}
   \min_{\X,\E}~\norm{\nabla(\X)}_{\circledast,\ell} &+\lambda\norm{\E}_1,\\
    \text { s.t.}~\Pomega(\X+\E)&= \Pomega(\Y). \\  
    \end{aligned}
\end{equation}
where $\norm{\nabla(\X)}_{\circledast,\ell}$ simultaneously encodes the global low-rankness and local consistency of tensorial traffic data, $\norm{\E}_1$ serves as a regularizer for sparse noise characterization, and $\lambda=1/\sqrt{max(n_1,n_2)*n_3}$  is universal \citep{hu2023flexible}, rendering the model parameter-free. Under the noise-aware decomposition strategy, the proposed RTC-GTNLN model not only performs imputation on the noisy incomplete observations, but also effectively separates the actual traffic data tensor from the sparse noise tensor , as visually demonstrated in Figure \ref{fig:noise_decom}.

Finally, we discuss the connections and differences between the proposed RTC-GTNLN model and other similar models from previous works. 

First, the element-wise $\ell_1$-$\ell_2$ norm is a powerful tool in compressed sensing and matrix recovery \citep{yin2015minimization, lou2018fast, li2024novel}. The proposed TNLN extends the application of the $\ell_1$-$\ell_2$ norm into tensor modeling. Compared with conventional nonconvex tensor nuclear norms \citep{chen2020nonconvex,gao2020robust,hu2023flexible}, TNLN does not require truncation coefficients or exponential parameters, making it more implementation-friendly.

Second, the Frobenius norm of the gradient tensor, widely known as \textit{Total Variation} (TV), has been extensively utilized to encode local consistency in diverse tensor data \citep{ling2021t,wang2018traffic,wang2023guaranteed}. This regularization is frequently combined with other nuclear norms to capture both global and local characteristics of traffic tensors \citep{chen2021scalable,zeng2025flexible}. 
In contrast, the proposed GTNLN serves as a single regularizer that enforces a low-rank constraint while preserving local consistency effects comparable to TV regularization.

Third, the nuclear norm of feature tensors (e.g., gradient tensors), as an advanced tool for representing the global low-rank and local consistency of tensors, has some limitations in current usage. 
These include insufficient accuracy caused by the use of convex nuclear norms \citep{peng2022exact,wang2023guaranteed} and the dependence on positional information \citep{shu2024low}.
The GTNLN we developed addresses these issues as a non-convex regularizer that requires no auxiliary data input. Notably, GTNLN is the first single non-convex regularizer proven to simultaneously capture the global low-rank structure and local smoothness mechanisms of traffic data, as revealed through the quantitative relationship between GTNLN and TV.

Fourth, although many existing robust tensor
completion models can achieve simultaneous imputation and denoising, their limitations in  feature  modeling and prior integration constrain recovery performance \citep{jiang2019robust, gao2020robust, hu2023flexible, song2020robust}. The proposed RTC-GTNLN model adopts advanced prior integration techniques, embedding global low-rank and local consistency seamlessly into the RTC framework, significantly improving recovery accuracy. The superiority of the proposed RTC-GTNLN model will be validated in subsequent experiments.

\section{Optimization Algorithm}
\label{sec:Algorithm}

This section presents  the optimization algorithm of the proposed RTC-GTNLN  model  for traffic data recovery.
We employ the widely-used \textit{alternating direction method of multipliers} (ADMM) \citep{boyd2011distributed,bai2018generalized,bai2022inexact} to solve the model \eqref{RTC-GTNLN}. First of all, we introduce a gradient tensor $\G= \nabla(\X) $ and an auxiliary variable $\K$  to compensate the missing entries of $\X$  using the following indicative function:
\begin{equation}
	\deltabm_{\K,\Omegabm} = \begin{cases}
	      0, & \Pomega(\K)=\0, \\
	      +\infty, & \text{otherwise}.	
		   \end{cases}
\end{equation}
Then the RTC-GTNLN model \eqref{RTC-GTNLN} can be rewritten as below:
\begin{equation}
    \begin{aligned}
    \min_{\X,\G,\K} \ & \norm{\G}_{\circledast,\ell} +\deltabm_{\K,\Omegabm}+\lambda \norm{\E}_1,\\
    \text { s.t.}~&\G=\nabla(\X),\\
    ~&\X+\E+\K= \Pomega(\Y). \\
    \end{aligned}
    \label{tlsv_au}
\end{equation}
Note that $\norm{\G}_{\circledast,\ell}=\sum_{i=1}^3 \alpha_i(\norm{\Gm_i}_*-\norm{\Gm_i}_F)$, where
$\Gm_i$ is the mode-$i$ unfolding matrix of $\G$. Then we introduce  auxiliaries variable $\Zm_i=\Gm_i$.
\begin{equation}
    \begin{aligned}
    \min_{\X,\G,\Zm_i,\K} \ & \sum_{i=1}^3 \alpha_i(\norm{\Zm_i}_*-\norm{\Zm_i}_F)+\deltabm_{\K,\Omegabm}+\lambda \norm{\E}_1,\\
    \text { s.t.}~&\G=\nabla(\X),\\
    ~&\X+\E+\K= \Pomega(\Y), \\
    ~&\Zm_i= \Gm_i. \\
    \end{aligned}
    \label{tlsv_auU}
\end{equation}
For model \eqref{tlsv_au}, we write its augmented Lagrangian function 
\begin{equation}
    \begin{aligned}
	& \ \ \ \ \ L(\X,\G,\K,\Zm_i,\E,\M,\N,\Qm_i) \\
   &=\sum_{i=1}^3 \alpha_i(\norm{\Zm_i}_*-\norm{\Zm_i}_F+\frac{\mu_t}{2}\norm{\Zm_i-\Gm_i+{\Qm_i}/{\mu_t}}_F^2)\\ 
  & + \frac{\mu_t}{2}\normlarge{\nabla(\X)-\G+{\M_k}/{\mu_t}}_F^2+\deltabm_{\K,\Omegabm}+\lambda \norm{\E}_1\\
 & +\frac{\mu_t}{2}\normlarge{\Pomega(\Y)-\X-\E-\K+{\N}/{\mu_t}}_F^2+\C,\\
    \end{aligned}
    \label{tlsv_lf}
\end{equation}
where $\M,\N$ and $\Zm_i(i=1,2,3)$ are Lagrange multipliers, $\mu_t>0$ is a positive scalar, and $\C$ is only the multipliers dependent squared items.

According to the augmented Lagrangian function, the ADMM optimization framework transforms the minimization problem for \eqref{tlsv_lf} into the following subproblems for each variable in an iterative manner, i.e.,
\begin{align}
\X^{t+1}&=\operatorname{arg}\min_{\X} ~L(\X,\G^t,\K^t,\Zm_i^t,\E^t,\M^t,\N^t,\Qm_i^t),\label{subb1} \\
\G^{t+1}&=\operatorname{arg}\min_{\G} ~L(\X^{t+1},\G,\K^t,\Zm_i^t,\E^t,\M^t,\N^t,\Qm_i^t),\label{subb2}\\
\K^{t+1}&=\operatorname{arg}\min_{\K} ~L(\X^{t+1},\G^{t+1},\K^t,\Zm_i^t,\E^t,\M^t,\N^t,\Qm_i^t),\label{subb3}\\
\Zm_i^{t+1}&=\operatorname{arg}\min_{\Zm_i} ~L(\X^{t+1},\G^{t+1},\K^{t+1},\Zm_i,\E^t,\M^t,\N^t,\Qm_i^t),\label{subb4}\\
\E^{t+1}&=\operatorname{arg}\min_{\E} ~L(\X^{t+1},\G^{t+1},\K^{t+1},\Zm_i^{t+1},\E,\M^t,\N^t,\Qm_i^t),\label{subb5}
\end{align}
and the multipliers is updated by
\begin{align}
\M_k^{t+1}&=\M_k^{t}+\mu_t(\nabla(\X^{t+1})-\G_k^{t+1}),\label{subb6}\\
\N^{t+1}&=\N^{t}+\mu_t(\Pomega(\Y)-\X^{t+1}-\E^{t+1}-\K^{t+1}),\label{subb7}\\
\Qm_i^{t+1}&=\Qm_i^{t}+\mu_t(\Zm_i^{t+1}-\Gm_i^{t+1}),\label{subb8}
\end{align}
where $t$ denotes the count of iteration in the ADMM. In the following, we deduce the solutions for  Updating $\X^{t+1}$, $\G^{t+1}$, $\K^{t+1}$,$\Zm_i$ and $\E$ respectively, each of which has the closed-form solution.

\textit{1) Updating $\X^{t+1}$}:
Take the derivative for augmented Lagrangian function (\ref{subb1}) with respect to $\X$ and set it to 0, then it gets the following linear equation for updating $\X^{t+1}$
\begin{equation}\label{tlsv_mj}
  \X^{t+1} + \nabla^T\nabla(\X^{t+1} )=\W^t,
\end{equation}
where $\nabla^T(\cdot)$ denotes the transpose operator of $\nabla(\cdot)$ and
\[\W^{t} =  \nabla^T(\G^t-\M^t/ {\mu_t}) +\Pomega(\Y)-\K^t-\E^t+\N^t/{\mu_t}.\]

Following lemma 1 in \cite{shu2024low}, we can obtain the closed-form solution \eqref{tlsv_mj}
\begin{equation}\label{x-updating}
   \X^{t+1} =\frac{\W^{t} \times_2 \Am^\top} {\1+ \1 \times_2 \Sm }\times_2 \Am,
\end{equation}
where $\mathbf{1}$ is a tensor with all entries as 1, the division is componentwise, the matrix $\Sm$ and $\Am$ are obtained by the spectral decomposition $\Dm^\top\Dm \Am = \Am \Sm$, the matrix $\Dm$ is the row circulant matrix associated with the gradient operator $\nabla(\cdot)$.

\textit{2) Updating $\G^{t+1}$}: Similarly, 
find the derivative of \eqref{subb2}  with respect to $\G$  and set it to 0, then we get the updating formula
\begin{equation}\label{g-updating}
   \G^{t+1} = (\sum_{i=1}^3 \operatorname{fold}_i(\Zm_i^t+\Qm_i^t/\mu_t)+\nabla(\X^{t+1})+\M^t/\mu_t)/4.
\end{equation}
where $\operatorname{fold}_{i}(\cdot)$  is the folding operator for converting a matrix to a third-order tensor along the $i$-th mode.

\textit{3) Updating $\K^{t+1}$:} 
The optimization of $\mathcal{K}$ obeys $\Pomega(\K) = \0$, which is updated by performing :
\begin{equation}\label{k-updating}
\K^{t+1} = \Pomega(\Y)-\X^{t+1}-\E^t+{\N^t}/{\mu_t}, \
\Pomega(\K^{t+1}) = \0.
\end{equation}

\textit{4) Updating $\Zm_i^{t+1} (i=1,2,3)$}: Fixed other variables in augmented Lagrangian function, then the subproblem \eqref{subb4} degenerates into 
\begin{equation}
    \begin{aligned}
   \operatorname{arg}\min_{\Zm_i} \frac{\alpha_i}{\mu_t}(\norm{\Zm_i}_{*}-\norm{\Zm_i}_F)+
 \frac{1}{2}\norm{\Zm_i-\Gm_i^{t+1}+\frac{\Qm_i^{t+1}}{\mu_t}}_F^2,
    \end{aligned}
\end{equation}
where $\Gm_i^{t+1}=\operatorname{unfold}_i(\G)$.
Using von Neumann's trace inequality \citep{lewis2005nonsmooth},
we can get a globally feasible solution 
\begin{equation}\label{z-updating}
\Zm_i^{t+1}=\Um_i\operatorname{diag}\left(\operatorname{prox}_{\frac{\alpha_i}{\mu_t},\ell_1-\ell_2}(\boldsymbol{\sigma}_i)\right)\Vm_i^\top
\end{equation}\label{NN_PRO}
where $\Um_i\operatorname{diag}(\boldsymbol{\sigma}_i)\boldsymbol{V}_i^{\top}$
is the singular value decomposition of $\Gm_i^{t+1}-\Qm_i^{t+1}/\mu_t$, 
$\boldsymbol{\sigma}_i$ is the singular value vector of $\Gm_i^{t+1}-\Qm_i^{t+1}/\mu_t$, and
$\operatorname{prox}_{\frac{\alpha_i}{\mu_t},\ell_1-\ell_2}(\boldsymbol{\sigma}_i)$ represents the proximity operator of  vector $\ell_1-\ell_2$ norm with parameter ${\alpha_i}/{\mu_t}$, i.e.,  
\begin{equation}\label{l_pro}
\operatorname{prox}_{\frac{\alpha_i}{\mu_t}}(\boldsymbol{\sigma}_i)=\operatorname{arg}\min_{\mathbf{y}} \frac{\alpha_i}{\mu_t}(\norm{\boldsymbol{\sigma}_i}_{1}-\norm{\boldsymbol{\sigma}_i}_2)+
 \frac{1}{2}\norm{\boldsymbol{\sigma}_i-\mathbf{y}}_2^2.
\end{equation}
Noted that \eqref{l_pro} can be easily solved according to \cite{li2024novel}.

\textit{5) Updating $\E^{t+1}$}: Fixing other variables
except for $\E$ in \eqref{subb5}, we can obtain the following sub-problem
\begin{equation}\label{1_pro}
\min_{\E}~\lambda \norm{\E}_1+\frac{\mu_t}{2}\normlarge{\Pomega(\Y)-\X-\E-\K+{\N}/{\mu_t}}_F^2.
\end{equation}
Then the closed-form solution of \eqref{1_pro} can be obtained by resorting to the element-wise shrinkage operator \cite{boyd2011distributed}, that is,
\begin{equation}\label{e-updating}
\E^{t+1}=\mathcal{S}_{\frac{\lambda}{\mu}}(\Pomega(\Y)-\X-\K+{\N}/{\mu_t})
\end{equation}
where $\mathcal{S}_\rho(\mathbf{x}):=\operatorname{sign}(\mathbf{x}) * \max \{|\mathbf{x}|-\rho, 0\}.$

\begin{algorithm}[tbp]\vspace{-1mm}
\renewcommand{\algorithmicrequire}{ \textbf{Input}:}
\renewcommand{\algorithmicensure}{ \textbf{Output}:}
\caption{ADMM for solving the RTC-GTNLN model \eqref{RTC-GTNLN}}\label{alg1}
\begin{algorithmic}[1]
\REQUIRE observed tensor $\Y$, weight coefficient $\alpha_i=1/3 (i=1,2,3)$, parameter $\lambda=1/\sqrt{max(n_1,n_2)*n_3}$.
\STATE Initialize $\X^0=\Pomega(\Y)$, $\G^0= \nabla (\X^0)$, $\Zm_i^0=\Qm_i^0=\mathbf{O}$, $\K^0=\E^0=\M^0=\N^0=\0$, $\mu_0=1e-6$.
\STATE \textbf{while} not converge \textbf{do}
\STATE \quad Update $\X^{t+1}$ by \eqref{x-updating};
\STATE \quad Update $\G^{t+1}$ by \eqref{g-updating};
\STATE \quad Update $\K^{t+1}$ by \eqref{k-updating};
\STATE \quad Update $\Zm_i^{t+1}$ by \eqref{z-updating};
\STATE \quad Update $\E^{t+1}$ by \eqref{e-updating};
\STATE \quad Update multipliers $\M^{t+1}$ by \eqref{subb6};
\STATE \quad Update multipliers $\N^{t+1}$ by \eqref{subb7};
\STATE \quad Update multipliers $\Qm_i^{t+1}$ by \eqref{subb8};
\STATE \quad Let $\mu_{t+1}=1.1\mu_t$; $t = t +1$.
\STATE \textbf{end while}
\ENSURE imputed tensor $\hat{\X}=\X^{t+1}$.
\end{algorithmic}
\end{algorithm}

The whole optimization procedure for solving the proposed RTC-GTNLN  model \eqref{RTC-GTNLN} is summarized in Algorithm \ref{alg1}.


Last but not least, we analyze the computational complexity of the proposed Algorithm \ref{alg1}. In each iteration, the update for $\X$ primarily consists of a mode product and a spectral decomposition, with corresponding time complexities of 
and $O(n_1n_3n_2^2)$ and $O(n_2^3)$, respectively. The computations for $\Zm_i$ involve singular value decompositions of unfolded matrices, each requiring a time complexity of $O(n_1n_2n_3(n_1+n_2+n_3))$, under the assumption that $n_i\leq\prod_{j\neq i}n_j$.
For other updates for $\K$, $\E$ and $\Qm_i$ , there only contains the componentwise operation with time complexity $O(n_1n_2n_3)$, and the update for $\G$ and $\M$  need further gradient operation with time complexity $O(n_1n_3n_2^2)$.
Overall, the per-iteration computational complexity of Algorithm \ref{alg1} is $O(n_1n_2n_3(n_1+n_2+n_3))$, assuming $n_i\leq\prod_{j\neq i}n_j$.

\section{Experiments}
\label{sec:experiments}
In this section, we perform comprehensive experiments to validate the effectiveness of our proposed RTC-GTNLN method. The evaluation is conducted on three widely recognized traffic datasets, all of which are standard benchmarks adopted in previous research. All the experiments are run on the following platforms: Windows 11 with 12th Gen Intel(R) Core(TM) i7-12700 2.1GHz CPU and 20GB memory. In the settings of software, we used Matlab (R2023b) and  Python 3.12.7 (with the Numpy and Pandas packages).

\subsection{Real-world Traffic Datasets}
\begin{enumerate}
    \item \textbf{PeMS04}: The PeMSD4 dataset, provided by California Performance of Transportation (PeMS), includes speed data from 307 detectors on 29 highways in the San Francisco Bay Area. Collected every 30 seconds and aggregated into 5-minute intervals, it spans January to February 2018. The final data structure can be constructed as a 3-way tensor of dimensions $307 \times 288 \times 59$.
    \item \textbf{PeMS08}: The PeMS08 dataset comprises traffic flow data collected over a two-month period (July to August 2016) in San Bernardino, California. The data was captured by 170 sensors distributed along 8 key roadways and aggregated by the California Performance of Transportation (PeMS). The final PeMSD8 tensor is structured as $170\times 288 \times 61$.
    \item \textbf{Guangzhou}: The Guangzhou dataset \footnote{\url{https://doi.org/10.5281/zenodo.1205229}} provides urban traffic speed measurements for 214 road sections in Guangzhou, China, spanning from August 1 to September 30, 2016, with a temporal resolution of 10 minutes. The dataset is represented as a tensor of size $214\times 144\times 61$.
\end{enumerate}

\subsection{Baseline Methods}
To evaluate the proposed RTC-GTNLN method, we compare it with the following baselines:
\begin{enumerate}

\item LRTC-TNN \citep{chen2020nonconvex}: This is a low-rank tensor completion method based on truncated nuclear norm minimization, capable of preserving key low-rank patterns in traffic  data.

\item LATC \citep{chen2021low}: This method combines  autoregression into low-rank tensor completion to enforce local consistency constraints in traffic data.

\item LRTC-3DST \citep{shu2024low}: A low-rank tensor completion method that fuses traffic-specific priors, effectively integrating low-rankness with three types of spatiotemporal features and thus avoids trade-off parameters.

\item RTC-SPN \citep{gao2020robust}: It's a robust tensor completion method that encodes the low-rankness of tensors using the Schatten $p$ norm. By separating low-rank and sparse components, it simultaneously performs data completion and denoising.

\item RTC-tubal \citep{jiang2019robust}: It  is a classical robust tensor completion model based on the tubal rank. It encodes the low-rank component of the target tensor using the tubal nuclear norm.

\item RTC-TTSVD \citep{song2020robust}: A robust low-rank tensor completion model within the T-SVD framework that replaces the standard T-SVD transform with a data-adaptive approach, making it more suitable for traffic-related data.

\item RTC-TCTV \citep{wang2023low}: TCTV is a recently proposed tool that combines low-rankness and smoothness with theoretical guarantee. Notably, it has been used for tensor completion (LRTC-TCTV) and denoising (RTPCA-TCTV) separately. For a fairer comparison, we integrate these completion and denoising models into a unified robust tensor completion model (RTC-TCTV).
\end{enumerate}

\subsection{Experiment Setup}

\subsubsection{Noisy pattern}
\begin{figure}[t]
\renewcommand{\arraystretch}{0.5}
\setlength\tabcolsep{0.5pt}
\centering
\begin{tabular}{ccccccc}
\centering
\includegraphics[width=29.3mm, height = 28.3mm]{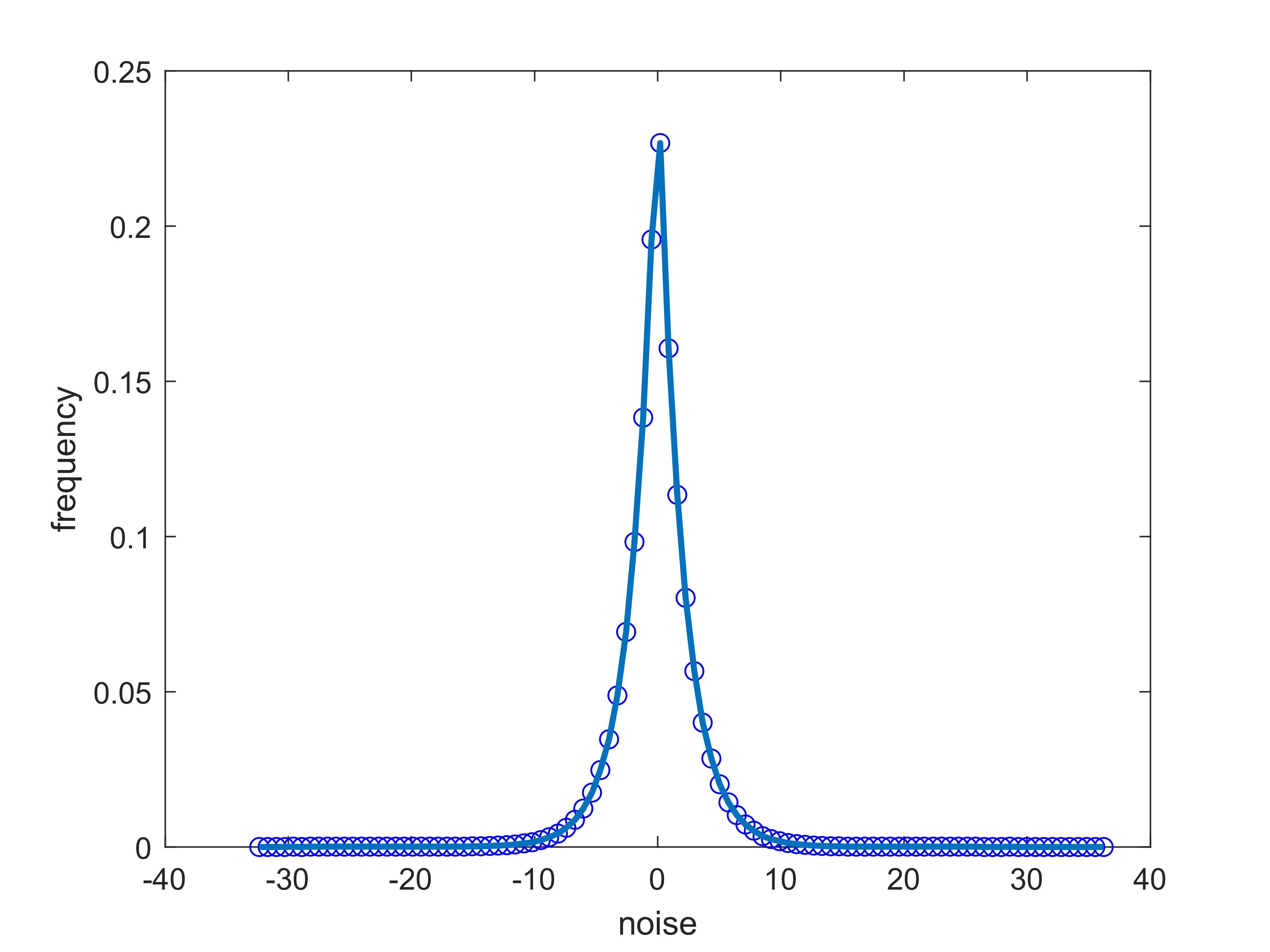}&
\includegraphics[width=29.3mm, height = 28.3mm]{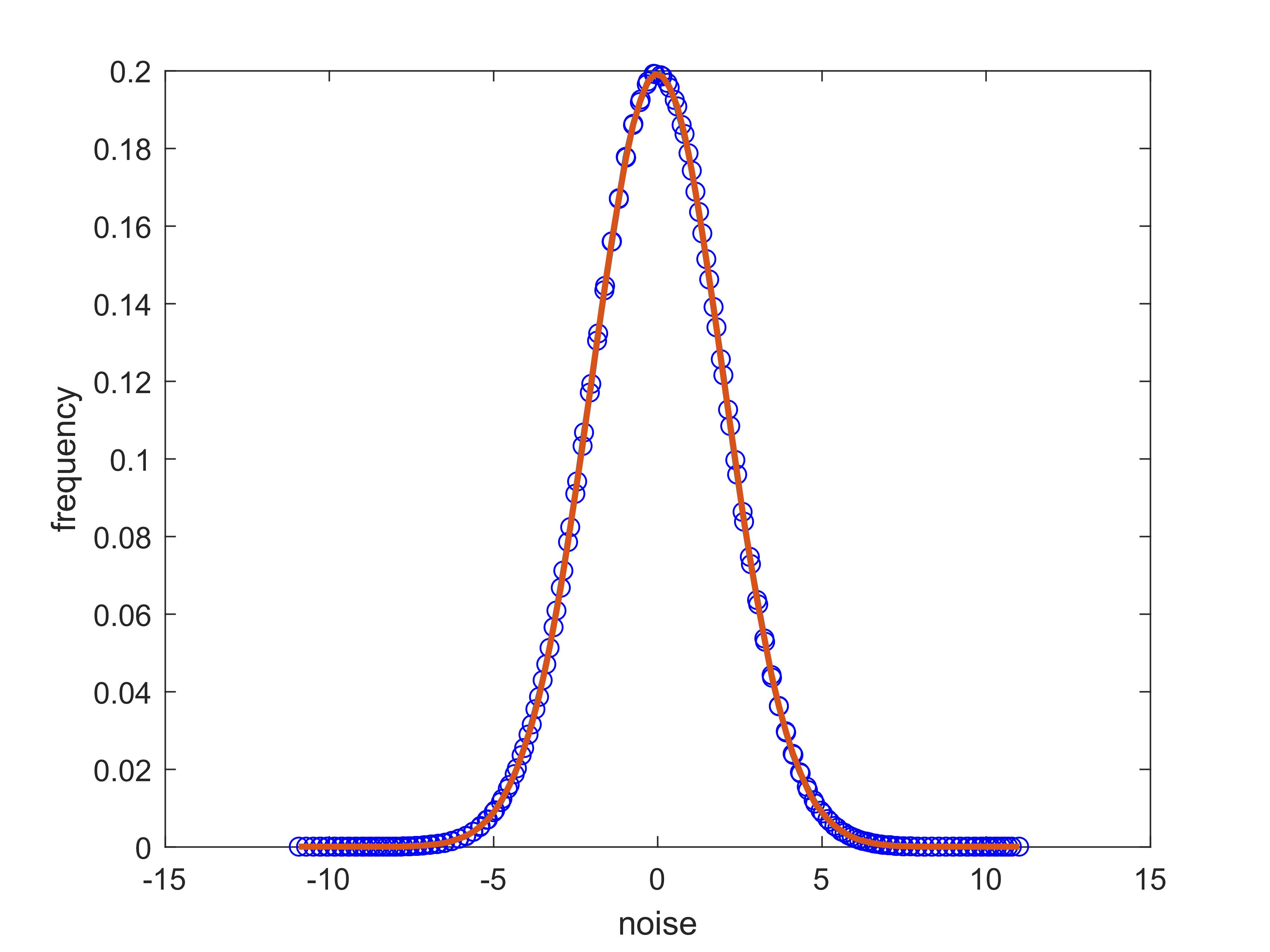}&
\includegraphics[width=29.3mm, height = 28.3mm]{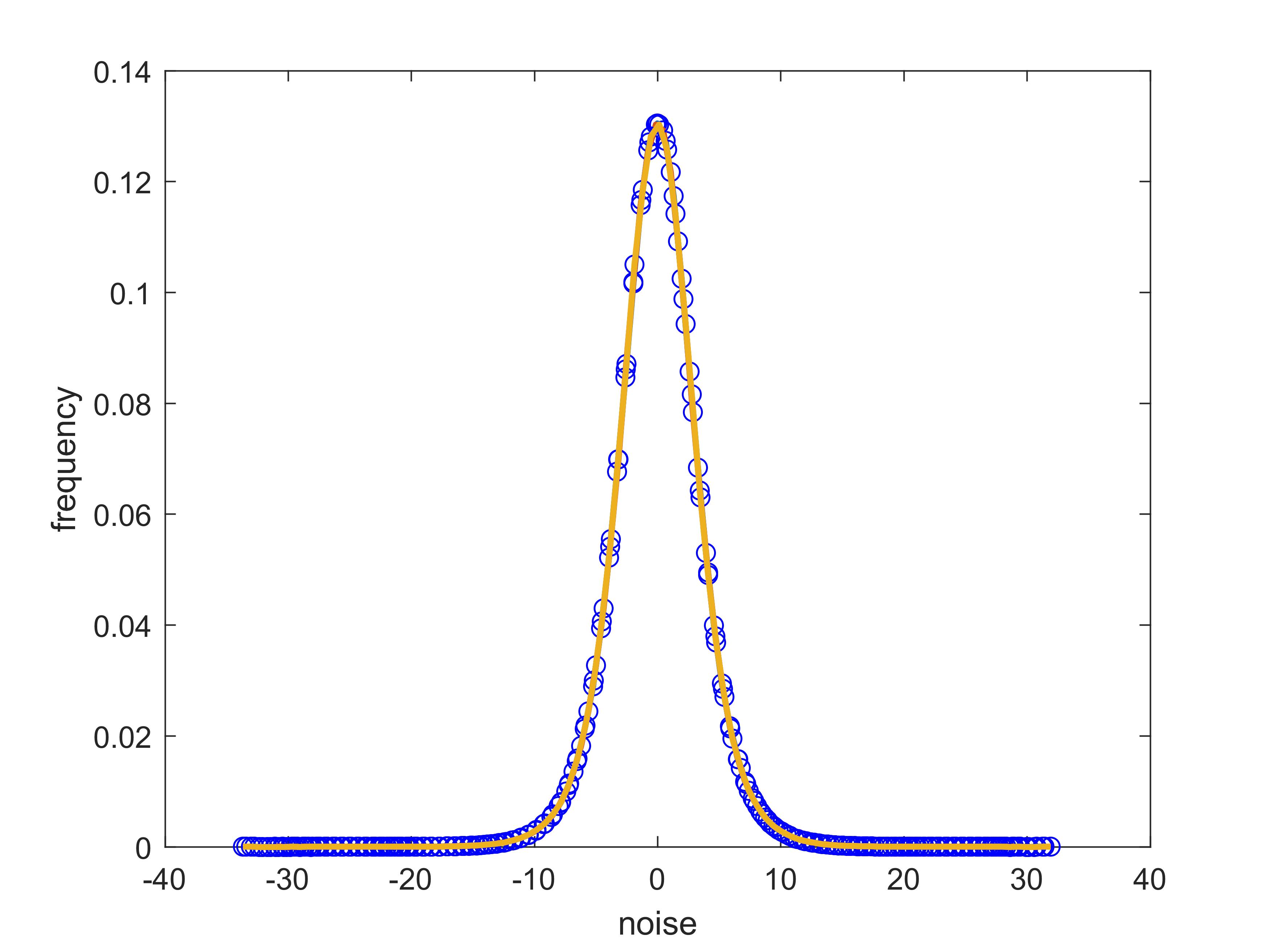}\\

 \scriptsize \textbf{Laplace Noise}& \scriptsize \textbf{Gaussian Noise} &\scriptsize \textbf{Mixed noise} \\

\scriptsize \textbf{(LN)}& \scriptsize \textbf{(GN)} &\scriptsize \textbf{(MN)} \\

\end{tabular}
\caption{Schematic representations of probability density functions for the three types of noise with mean $\mu=0$.
}\label{fig.noisy pattern}
\end{figure}

We introduce three common types of  noise  encountered in traffic data collection and transmission:
\begin{enumerate}
  \item Laplace Noise (LN): The probability density function of Laplace noise is defined as:
  \[
   f_{Laplace}(x \mid \mu, b)=\frac{1}{2 b} \exp \left(-\frac{|x-\mu|}{b}\right),
\]
 where $\mu$ is the mean, and $b>0$ is the scale parameter. The Laplace distribution has a sharp peak around the mean and heavier tails than the Gaussian distribution.
 In traffic data, Laplace noise can simulate sudden deviations due to abnormal events, such as accidents, extreme weather, and traffic signal failures.
  \item Gaussian Noise (GN): The PDF of Gaussian noise is given by:
  \[
f_{Gaussian}\left(x \mid \mu, \sigma^2\right)=\frac{1}{\sqrt{2 \pi \sigma^2}} \exp \left(-\frac{(x-\mu)^2}{2 \sigma^2}\right),
\]
where $\mu$ is the mean, and $\sigma>0$ is the standard deviation. The Gaussian distribution has a bell-shaped curve, with the highest probability density near the mean and faster decay in the tails compared to the Laplace distribution. Gaussian noise can simulate non-systematic errors caused by insufficient sensor accuracy and environmental changes.
  \item Composite Noise (CN): 
  When two independent noise sources (Gaussian and Laplace) are combined, the probability density function of the resulting noise is the convolution of the original two distributions. This combination describes the physical superposition of two types of noise. For example, during signal transmission, the signal may be simultaneously affected by sensor errors (Gaussian noise) and sudden abnormal interference (Laplace noise).
  The probability density function of composite noise can be obtained using the convolution formula:
  \[ f_Z(z) = \int_{-\infty}^{\infty} f_{\text{Gaussian}}(x) f_{\text{Laplace}}(z-x) dx, \]
where $Z = Y_1 + Y_2$ ,   $Y_1$ and  $Y_2$  are the Gaussian and Laplace noise sources, respectively.
\end{enumerate}
 The schematic representations of the probability density functions for these three noise categories are shown in Fig.\ref{fig.noisy pattern}.
In subsequent experiments, we configure the noise parameters as follows: For Laplace noise simulation, we configure the parameters as $b=3$ and $b=5$ to generate point-wise Laplace noise instances LN-1 and LN-2 respectively. In Gaussian noise modeling, we select standard deviation values of $\sigma=3$ and $\sigma=5$ to simulate point-wise Gaussian noise instances GN-1 and GN-2. For composite noise generation, we employ parameter pairs $(b,\sigma)=(2,2)$ and $(3,3)$ to characterize point-wise composite noise instances CN-1 and CN-2 correspondingly. Throughout these configurations, all noise distributions maintain zero mean values ($\mu=0$).



\begin{figure}[t]
	\centering
		\includegraphics[width=0.9\linewidth]{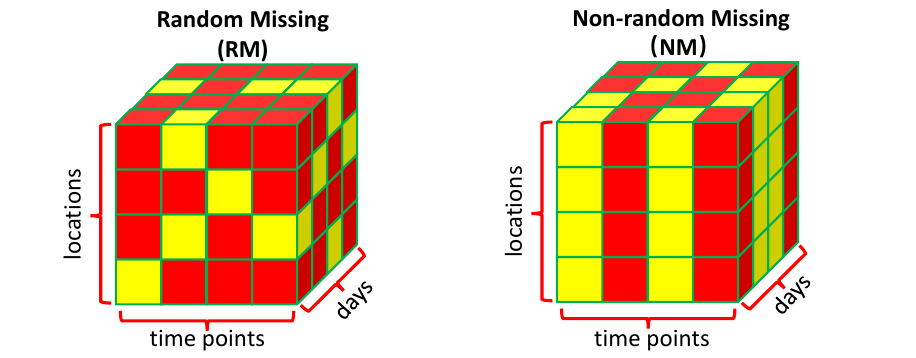}
            \centering
	\caption{Tensor diagrams illustrating different missing pattern. The red areas are observed and the yellow are missing.}
	\label{RNM}
\end{figure}

\subsubsection{Missing pattern}
We adopt two types of data missing patterns: one being the pointwise \textit{Random Missing} (RM) mechanism where the sampling set follows a Bernoulli distribution 
$\Omega\sim\operatorname{Ber}(p)$ with  $p$ as the sampling probability, and the other being the prevalent \textit{Non-random Missing} (NM) pattern – specifically, fiber-like missing structures that characterize regional missing configurations emerging from consecutive sensor failures induced by power shortages or malfunctions in internet data centers. 
Schematic representations of these missing patterns are demonstrated in Fig.\ref{RNM}.




\subsubsection{Evaluation criteria}
We use the MAE and RMSE to assess the recovery performance, defined by
\begin{equation}
    \text{MAE}=\frac{1}{n_1n_2n_3}\sum_{i,j,k}\left|(\X_0)_{ijk}-\hat{\X}_{ijk}\right|,
\end{equation}
\begin{equation}
    \text{RMSE}=\sqrt{\frac{1}{n_1n_2n_3}\sum_{i,j,k}\left((\X_0)_{ijk}-\hat{\X}_{ijk}\right)^2},
\end{equation}
where  $\X_0$ and $\hat{\X}$ are the actual traffic tensor data and corresponding recovery result, respectively,  The lower value of MAE and RMSE indicates more precise recovery performance.

\begin{table*}[t]
\renewcommand{\arraystretch}{1}
\centering
\caption{Performance comparison (in MAE/RMSE) of LRTC-GTNLN and baseline models with various noisy patterns  for traffic data recovery from noisy incomplete observations. The degradation cases column indicates the noisy pattern and the proportion of noise/missing values.}
\vspace{-0.2cm}
\label{noiseee}
\centering
\footnotesize
\resizebox{\textwidth}{!}{ 
\begin{tabular}{cc|cccccccc}
\toprule
Dataset&Degradation cases&LRTC-TNN&LATC&LRTC-3DST&RTC-SPN&RTC-tubal&RTC-TTSVD&RTC-TCTV&RTC-GTNLN \\
\midrule
\multirow{6}{*}{\textbf{PeMS04}}

& 50\%LN-1+50\%missing& 2.38/3.76  & 2.29/\underline{3.46}&  2.61/3.71&2.39/3.78& 
2.29/3.86&  2.07/3.66& \underline{1.94}/3.50& \textbf{1.68}/\textbf{2.55} \\
& 50\%LN-2+50\%missing & 3.49/5.56& 3.55/5.44 &  4.14/5.91&3.52/5.64& 
3.09/4.55&  2.69/4.14& \underline{2.39}/\underline{3.91}& \textbf{2.29}/\textbf{3.18}\\

& 50\%GN-1+50\%missing& 2.02/3.03& 1.90/\underline{2.64} &   2.11/2.76& 2.00/2.95& 2.15/3.73& 1.98/3.57& \underline{1.88}/3.43& \textbf{1.63}/\textbf{2.46} \\
& 50\%GN-2+50\%missing& 2.90/4.22 & 2.86/3.98 &  3.30/4.28&2.92/4.26& 
2.90/4.31&  2.55/3.98& \underline{2.42}/\underline{3.81}& \textbf{2.24}/\textbf{3.04}\\

& 50\%CN-1+50\%missing& 2.17/3.29 & 2.06/\underline{2.94}&   2.31/3.11& 2.16/3.27& 2.22/3.78& 2.04/3.61& \underline{1.92}/3.48& \textbf{1.67}/\textbf{2.51} \\
& 50\%CN-2+50\%missing& 2.90/4.34 & 2.86/4.11&  3.30/4.44&2.93/4.42& 
2.81/4.25&  2.52/3.97& \underline{2.35}/\underline{3.76}& \textbf{2.14}/\textbf{2.96}\\
\midrule

\multirow{6}{*}{\textbf{PeMS08}}

& 50\%LN-1+50\%missing& 2.24/3.56& 2.20/\underline{3.38} &  2.61/3.69&2.24/3.59& 
1.98/3.72&  1.75/3.52& \underline{1.70}/3.46& \textbf{1.50}/\textbf{2.34}\\
& 50\%LN-2+50\%missing& 3.37/5.43& 3.46/5.35 &  4.20/5.95&3.39/5.53& 
2.62/4.17&  2.22/3.80& \underline{2.13}/\underline{3.71}& \textbf{2.04}/\textbf{2.88}\\

& 50\%GN-1+50\%missing& 1.88/2.79& 1.80/\underline{2.54}&   2.10/2.72& 1.88/2.75& 1.91/3.66& 1.73/3.48& \underline{1.68}/3.42& \textbf{1.47}/\textbf{2.29}\\
& 50\%GN-2+50\%missing& 2.77/4.05& 2.76/3.89&  3.33/4.29&2.78/4.12& 
2.53/4.07&  2.19/3.75&\underline{2.11}/\underline{3.67}& \textbf{2.03}/\textbf{2.80}\\

&50\%CN-1+50\%missing& 2.03/3.08 & 1.96/\underline{2.85}&   2.30/3.08& 2.03/3.06& 1.96/3.70& 1.76/3.50& \underline{1.71}/3.45& \textbf{1.50}/\textbf{2.32}\\
&50\%CN-2+50\%missing& 2.77/4.19& 2.77/4.04 &  3.33/4.44&2.78/4.26& 
2.44/4.02&  2.15/3.72& \underline{2.05}/\underline{3.64}& \textbf{1.93}/\textbf{2.72}\\
\midrule

\multirow{6}{*}{\textbf{Guangzhou}}

& 50\%LN-1+50\%missing & 2.72/3.94& 2.67/3.80 &  2.90/4.01&2.78/4.04& 
2.62/3.88&  \underline{2.39}/\underline{3.63}& 2.42/3.68& \textbf{2.31}/\textbf{3.35}\\
& 50\%LN-2+50\%missing & 3.82/5.68& 3.89/5.68 &  4.37/6.11&3.88/5.77& 
3.23/4.48&  2.83/4.03& \underline{2.80}/\underline{4.02}& \textbf{2.74}/\textbf{3.78}\\

& 50\%GN-1+50\%missing & 2.38/3.27 & \underline{2.28}/\underline{3.05}&   2.42/3.17& 2.42/3.35& 2.50/3.75& 2.33/3.56& 2.35/3.59& \textbf{2.24}/\textbf{3.25}\\
& 50\%GN-2+50\%missing & 3.24/4.38 & 3.23/4.29&  3.55/4.55&3.30/4.48& 
3.08/4.27&  \underline{2.74}/\underline{3.89}& 2.74/3.91& \textbf{2.67}/\textbf{3.66}\\

& 50\%CN-1+50\%missing& 2.52/3.51 & 2.44/\underline{3.32} &   2.61/3.47& 2.57/3.60& 2.56/3.82& \underline{2.35}/3.58& 2.39/3.63& \textbf{2.28}/\textbf{3.30}\\
& 50\%CN-2+50\%missing& 3.24/4.50& 3.23/4.41 &  3.56/4.71&3.29/4.60& 
3.01/4.21&  2.70/\underline{3.88}& \underline{2.69}/3.88& \textbf{2.61}/\textbf{3.61}\\
\bottomrule
\multicolumn{10}{l}{{The best results are highlighted with \textbf{bold} and the second best are highlighted with \underline{underline}}.}
\end{tabular}}
\end{table*}

\subsection{Experimental Results}
\subsubsection{The recovery results with various noisy patterns from noisy  
incomplete observations} 
To evaluate the capability of our proposed RTC-GTNLN model in simultaneously handling missing values and various types of noise, we set the data to have 50\% random missing values, while the remaining 50\% is affected by different types of noise. This resulted in six data degradation scenarios: (1) 50\% random  missing values + the remaining 50\% affected by Laplace noise 1 (LN-1) , (2) 50\% random missing values + Laplace noise 2 (LN-2), (3) 50\% random missing values + the remaining 50\% affected by  Gaussian noise 1 (GN-1), (4) 50\% random missing values +  Gaussian noise 2 (GN-2), (5) 50\% random missing values + composite noise 1 (CN-1), and (6) 50\% random missing values + composite noise 2 (CN-2). We conduct detailed experiments on these six degradation scenarios across three datasets to evaluate the performance of the proposed RTC-GTNLN model and other baseline methods. 

\begin{table}[h]
    \centering
\renewcommand{\arraystretch}{1}
\setlength{\tabcolsep}{3pt}
    \caption{Summary of the running time (in seconds) of different methods for recovery tasks across various datasets.}\label{time}
    \vspace{-0.1cm}
    \resizebox{0.3\textwidth}{!}{
    \begin{tabular}{c|c c c c c c c c}
        \hline
        Method/Dataset    & PeMS04 & PeMS08 & GuangZhou \\
        \hline
        LRTC-TNN &69& 37& 25\\
        LATC & 1096& 613& 447\\
        LRTC-3DST & 222& 109& 73\\
        RTC-SPN & 73& 28& 19\\
        RTC-tubal &98& 57& 36\\
        RTC-TTSVD &142& 63& 38\\
        RTC-TCTV &475& 255& 166\\
        RTC-GTNLN &102& 54& 33\\
        \hline
    \end{tabular}}
    \vspace{-0.3cm}
\end{table}

\begin{figure}[h]
\centering
\includegraphics[width=1\linewidth]{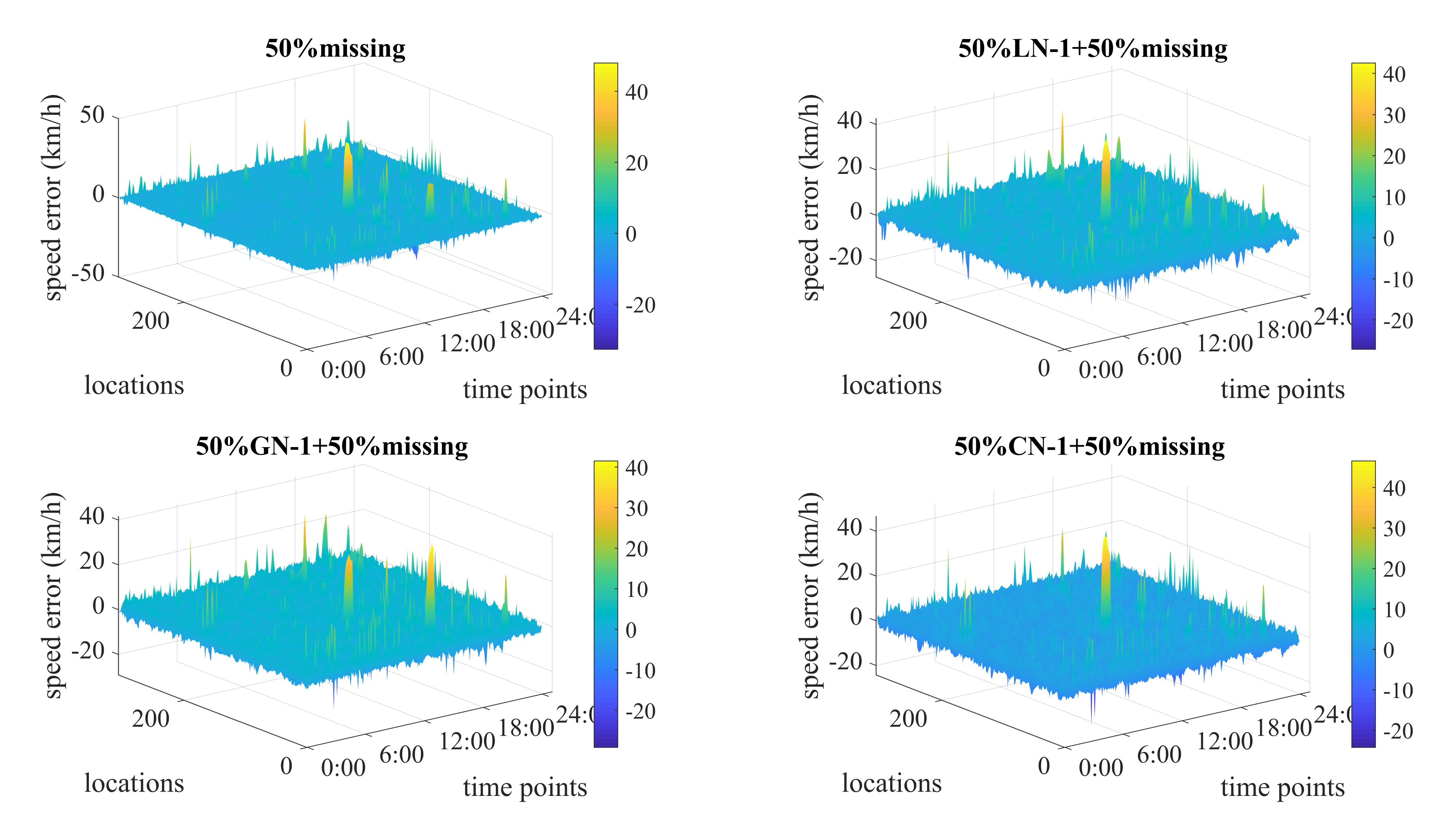}
\caption{The difference between the recovered values generated by the RTC-GTNLN model and the actual observed values on the 35th day, across various noise types in the PeMS04 traffic dataset}\label{fig.day35}
\vspace{-0.4cm}
\end{figure}

\begin{figure}[h]
\centering
\setlength{\belowcaptionskip}{-0.0cm}  
\includegraphics[width=1\linewidth]{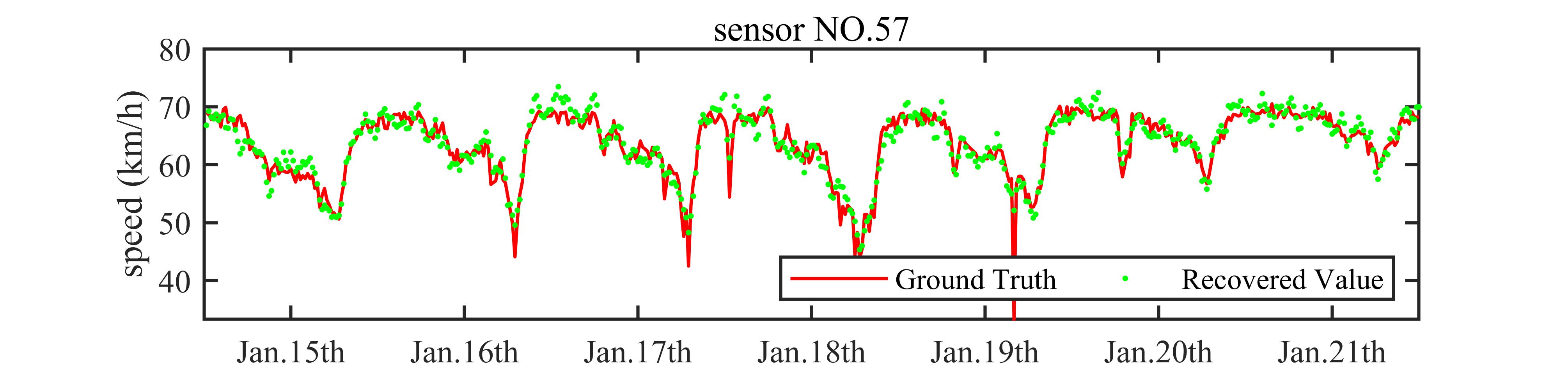}\\
\includegraphics[width=1\linewidth]{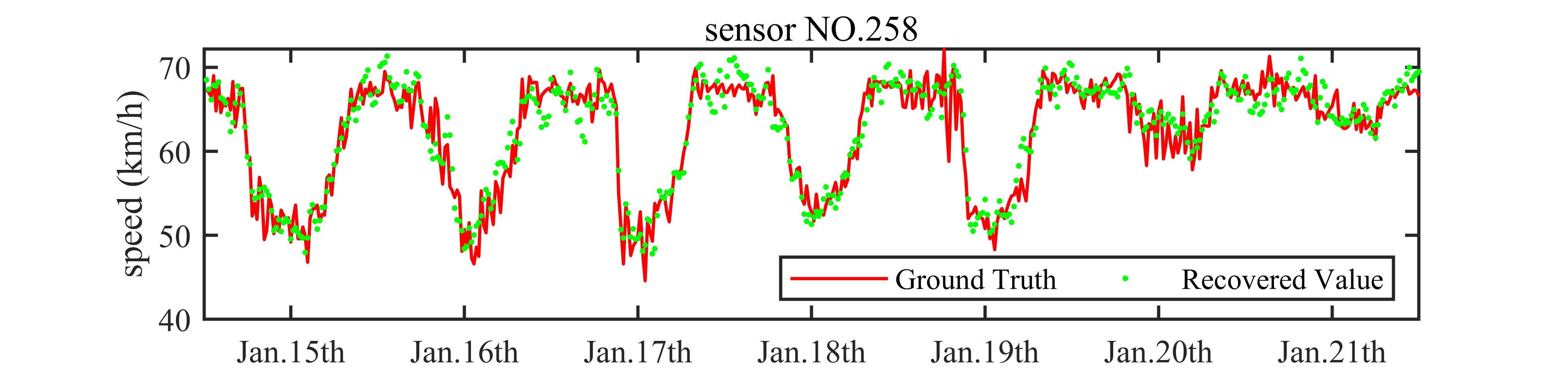}\\
   \footnotesize  \textbf{(a) PeMS04} \\  
\includegraphics[width=1\linewidth]{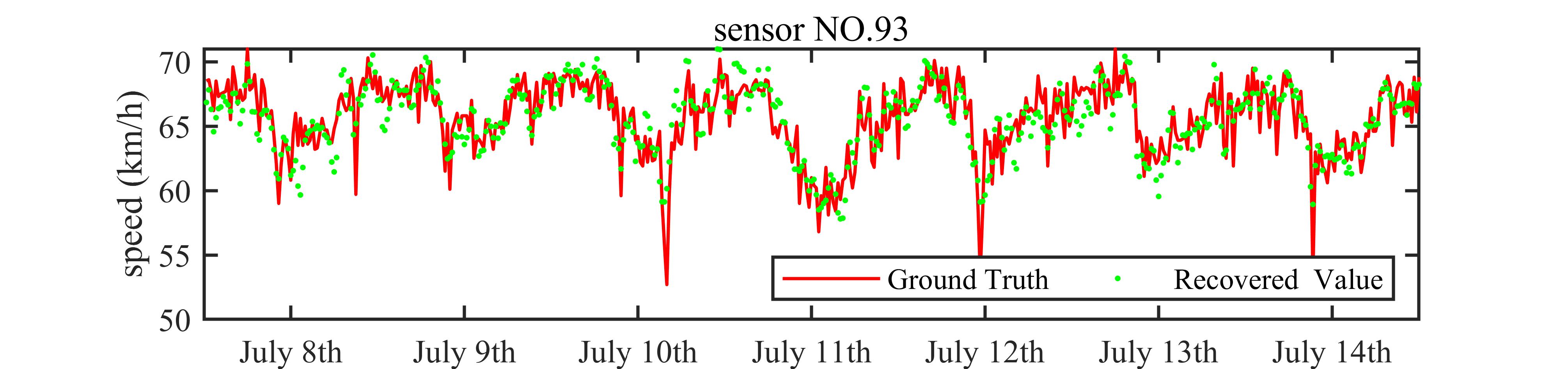}\\
\includegraphics[width=1\linewidth]{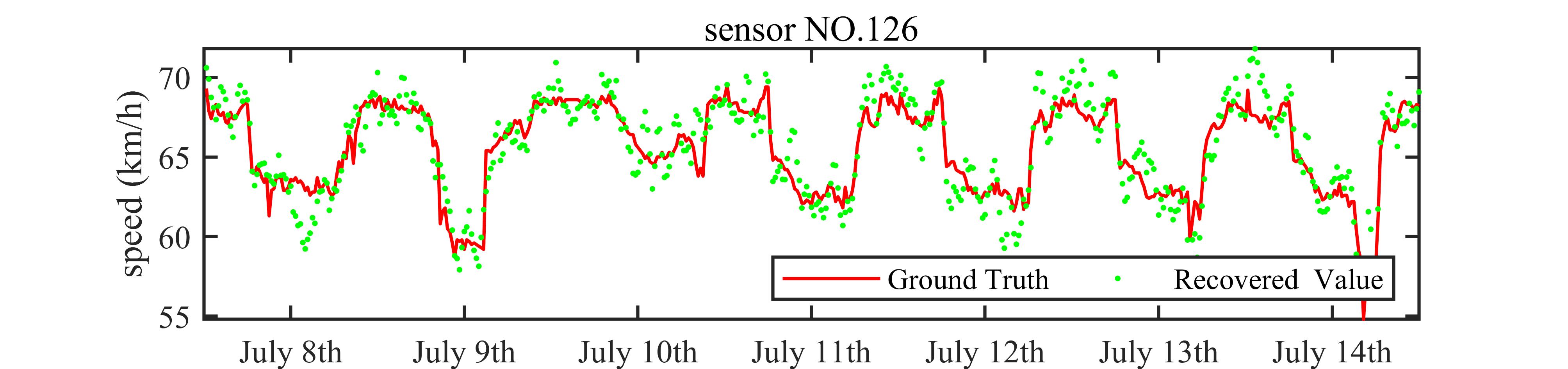}\\
     \footnotesize \textbf{(b) PeMS08}\\
\includegraphics[width=1\linewidth]{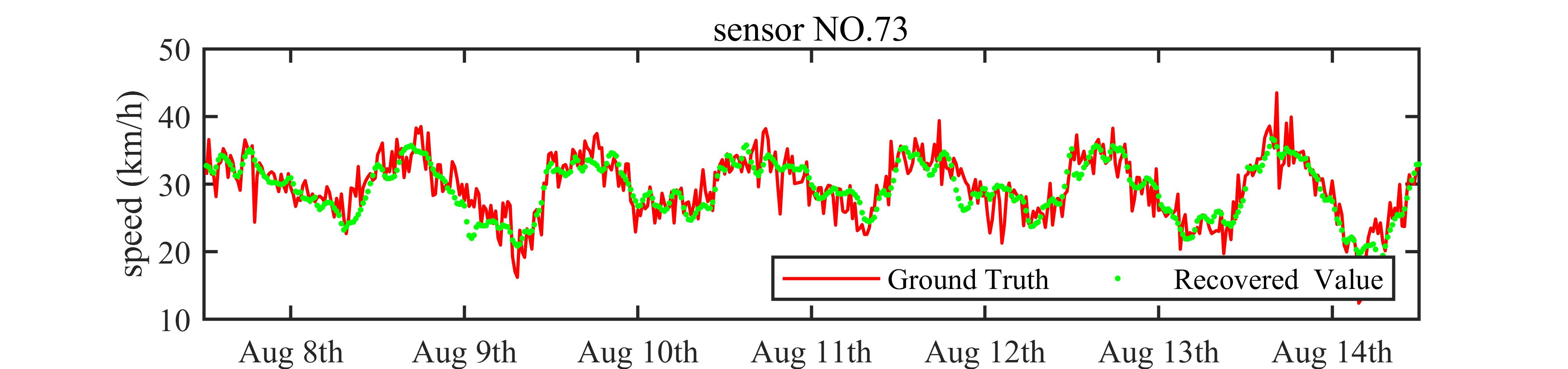}\\ 
\includegraphics[width=1\linewidth]{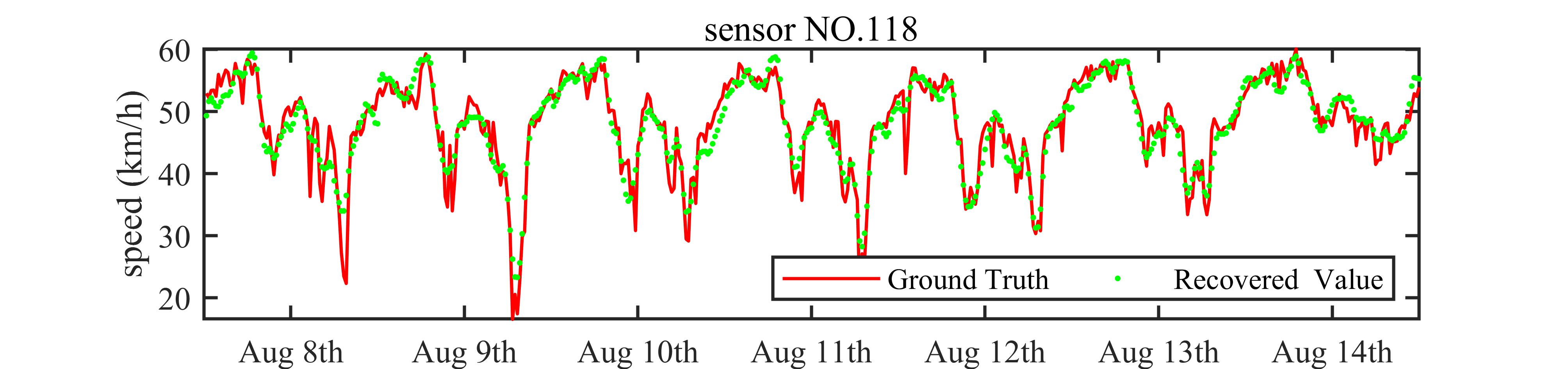}\\ 
     \footnotesize \textbf{(c) Guangzhou} \\ 
     \vspace{-0.1cm}
\caption{The recovered results obtained by RTC-GTNLN model for three traffic data sets. In these panels, red curves are the ground truth, while green points are the recovery values.}
\vspace{-0.5cm}
\label{04G_CN1_50}
\end{figure}

Table \ref{noiseee} summarizes the MAE and RMSE results of all methods under these degradation scenarios, considering 50\% random missing data combined with different types of noise across various datasets. 
Additionally, Table \ref{time} provides a comparison of the running time of different methods under the 50\% LN-1 + 50\% random missing degradation scenario across the datasets. As shown in Table \ref{noiseee}, the proposed RTC-GTNLN method significantly outperforms other baseline methods when simultaneously handling missing values and various types of noise, achieving the smallest MAE and RMSE values. In the case of purely missing data degradation, the recovery accuracy of the proposed RTC-GTNLN method is slightly lower but very close to that of the LRTC-3DST model. 
In addition, low-rank tensor completion methods such as LATC and LRTC-3DST have poor recovery results for strong noise situations (such as LN-2, GN-2, CN-2), while robust tensor completion methods can handle it perfectly, which is consistent with our previous discussion. From Table \ref{time}, we observe that high-accuracy recovery methods such as LATC, LRTC-3DST, and RTC-TCTV require significantly more computational time compared to the proposed RTC-GTNLN model.



To characterize the overall trend of differences between the recovered data and the original traffic data obtained by the proposed RTC-GTNLN method, we calculate the difference between the recovered values and the ground truth for the 35th day under different types of noise. The results are shown in Figure \ref{fig.day35}. From the figure, we can see that the differences between the recovered and ground truth values are very small across all locations under various noise conditions.
Furthermore, we randomly select specific examples from the 50\% LN-1 + 50\% random missing degradation scenario for the three datasets to analyze the performance of the proposed RTC-GTNLN model. As shown in Figure \ref{04G_CN1_50}, the recovered values obtained by the proposed RTC-GTNLN method are consistently close to the ground truth curve across all experiments in the datasets.

\subsubsection{The recovery results with various missing patterns
from  noisy incomplete observations} 
To evaluate the recovery performance of our proposed RTC-GTNLN model under different mixtures of noise and missing data, we conduct experiments with varying random missing  rates (20\%RM, 40\%RM, 60\%RM, 80\%RM) and non-random missing  rates (30\%NM, 50\%NM, 70\%NM). All observed entries were further corrupted with Laplace noise 1. For example, under a 20\% random missing rate, the remaining 80\% of the observed values were entirely contaminated by Laplace noise. Table \ref{different_misssing_pattern} summarizes the MAE and RMSE results of the proposed RTC-GTNLN and other baseline methods across various datasets under different random and non-random missing rates with Laplace noise.
According to the data in Table \ref{different_misssing_pattern}, our proposed RTC-GTNLN model achieves the lowest MAE and RMSE values under almost all missing rates. An interesting observation is that the accuracy of methods such as RTC-SPN, RTC-tubal, RTC-TTSVD, RTC-TCTV, and RTC-GTNLN decreases as the missing rate increases, whereas the accuracy of LRTC-TNN, LATC, and LRTC-3DST improves with higher missing rates. This phenomenon occurs because LRTC-TNN, LATC, and LRTC-3DST lack denoising capabilities; as the missing rate increases, the proportion of noise decreases, leading to improved recovery accuracy. \\
\indent To visualize the advantages of our proposed method, we test all methods on the PeMS08 dataset under the degradation scenario comprising 40\% random missing values and 60\% Laplace noise 1. 
The recovery results of the RTC-GTNLN model and baseline models for the 56th day of the PeMS08 dataset are shown in Figure \ref{pems08-all-methods}, along with residual plots obtained by subtracting the recovery outputs from the actual traffic data.  It is clear that the recovery results of LRTC-TNN, LATC, and LRTC-3DST are heavily contaminated with noise. While methods like RTC-SPN, RTC-tubal, RTC-TTSVD, and RTC-TCTV are capable of denoising, they fail to restore the fine details of the traffic data effectively. 
The visualization results demonstrate that the proposed RTC-GTNLN algorithm excels in both noise removal and detail preservation, with its residual plot exhibiting the cleanest map  among comparative methods.

\begin{table*}[!htbp]
\renewcommand{\arraystretch}{1}
\centering
\caption{Performance comparison (in MAE/RMSE) of LRTC-GTNLN and baseline models with various missing patterns for traffic data recovery from noisy incomplete observationss. The degradation cases column indicates the missing pattern and the proportion of missing values/noise.}

\label{different_misssing_pattern}
\centering
\footnotesize
\resizebox{\textwidth}{!}{ 
\begin{tabular}{cc|ccccccccc}
\toprule
Dataset&Degradation cases &LRTC-TNN& LATC &LRTC-3DST&RTC-SPN&RTC-tubal&RTC-TTSVD&RTC-TCTV&RTC-GTNLN\\
\midrule
\multirow{7}{*}{\textbf{PeMS04}}
& 20\%RM+80\%noise  & 2.72/4.00 & 2.69/3.91& 2.81/3.99&   2.72/3.99& 2.04/3.40& 1.94/3.30& \underline{1.68}/\underline{3.10}&\textbf{1.51}/\textbf{2.16}\\
& 40\%RM+60\%noise & 2.48/3.82& 2.41/3.61 & 2.67/3.78& 2.48/3.82& 2.19/3.69& 2.02/3.52& \underline{1.84}/\underline{3.34}& \textbf{1.61}/\textbf{2.39}\\
& 60\%RM+40\%noise & 2.29/3.73& 2.19/\underline{3.35}&  2.56/3.65&2.31/3.77& 
2.39/4.07&  2.12/3.80& \underline{2.06}/3.68& \textbf{1.78}/\textbf{2.75}\\
& 80\%RM+20\%noise & 2.21/3.86& \textbf{2.07}/\textbf{3.38}&  2.50/3.69&2.30/4.07& 
2.68/4.67&  2.28/4.22& 2.44/4.27& \underline{2.13}/\underline{3.40}\\
& 30\%NM+70\%noise &2.62/3.97 &2.54/3.75 &2.74/3.89 &2.32/4.29 &2.16/3.63 &1.99/3.42 &\underline{1.75}/\underline{3.22 }&\textbf{1.56/2.26 }\\
& 50\%NM+50\%noise&2.46/3.97 &2.29/\underline{3.46} &2.61/3.70 &2.44/4.65 &2.40/4.04 &2.10/3.69 &\underline{1.94}/3.54 &\textbf{1.69/2.56} \\
& 70\%NM+30\%noise&2.69/5.35 &\underline{2.12}/\underline{3.33} &2.52/3.64 &2.69/5.11 &2.78/4.79 &2.26/4.10 &2.24/4.05 &\textbf{1.91/3.01} \\
\midrule

\multirow{7}{*}{\textbf{PeMS08}}
& 20\%RM+80\%noise & 2.67/3.94 & 2.65/3.89 & 2.81/3.99&   2.69/3.91& 1.79/3.41& 1.66/3.25& \underline{1.51}/\underline{3.18}&\textbf{1.34}/\textbf{1.99}\\
& 40\%RM+60\%noise& 2.37/3.68& 2.34/3.54&   2.67/3.78& 2.63/3.74& 1.91/3.60& 1.73/3.42& \underline{1.63}/\underline{3.35}& \textbf{1.44}/\textbf{2.21}\\
& 60\%RM+40\%noise & 2.12/3.48 &2.06/\underline{3.22} &  2.56/3.62&3.14/4.22& 
2.06/3.86&  1.82/3.63& 1.80/3.59& \textbf{1.58}/\textbf{2.52}\\
& 80\%RM+20\%noise &  1.96/3.47& \textbf{1.86}/\underline{3.12}&  2.51/3.62&3.51/4.66& 
2.26/4.23&  1.95/3.95& 2.09/3.97& \underline{1.89}/\textbf{3.04}\\
& 30\%NM+70\%noise &2.54/3.88 &2.50/3.72 &2.73/3.88 &1.75/3.98 &1.86/3.52 &1.70/3.36 &\underline{1.56}/\underline{3.28} &\textbf{1.39/2.09 }\\
& 50\%NM+50\%noise&2.29/3.74 &2.21/\underline{3.39} &2.61/3.69 &1.92/4.28 &2.01/3.77 &1.80/3.54 & \underline{1.69}/3.48&\textbf{1.49/2.33} \\
& 70\%NM+30\%noise&2.46/5.40 &1.96/\underline{3.14} &2.75/4.14 &2.18/4.72 &2.22/4.10 &1.91/3.82 &\underline{1.90}/3.78 &\textbf{1.68/2.71} \\
\midrule

\multirow{7}{*}{\textbf{Guangzhou}}
& 20\%RM+80\%noise & 2.86/4.08& 2.84/4.03 & 2.92/4.09&   2.85/4.04& 2.30/3.43& \underline{2.17}/\underline{3.28}& 2.18/3.34&\textbf{2.03}/\textbf{2.89}\\
& 40\%RM+60\%noise& 2.75/3.97& 2.71/3.86&   2.89/4.02& 2.76/3.90& 2.50/3.72& \underline{2.31}/\underline{3.50}& 2.32/3.53& \textbf{2.20}/\textbf{3.18}\\
& 60\%RM+40\%noise & 2.69/3.93 & 2.64/3.78 &  2.91/4.02&2.86/4.14& 
2.76/4.09&  \underline{2.49}/\underline{3.78}& 2.54/3.83& \textbf{2.44}/\textbf{3.54}\\
& 80\%RM+20\%noise& 2.73/4.03& \textbf{2.66}/\textbf{3.86}&  3.00/4.19&3.62/4.85& 
3.13/4.63&  \underline{2.76}/4.19& 2.89/4.31& 2.86/\underline{4.14}\\
& 30\%NM+70\%noise &2.85/4.10 &2.77/3.94 &2.91/4.06 &2.49/3.83 &2.46/3.67 &2.27/\underline{3.43} &\underline{2.25}/3.44 &\textbf{2.11/3.02} \\
& 50\%NM+50\%noise&2.82/4.10 &2.69/3.82 &2.90/4.02 &2.73/4.16 &2.74/4.07 &\underline{2.42}/\underline{3.66} &2.46/3.76 & \textbf{2.32/3.36}\\
& 70\%NM+30\%noise&3.10/5.01 &\underline{2.67}/\underline{3.83} &2.96/4.10 &3.09/4.63 &3.17/4.68 &2.70/4.05 &2.76/4.19 &\textbf{2.62/3.82 }\\
\bottomrule
\multicolumn{10}{l}{{The best results are highlighted with \textbf{bold} and the second best are highlighted with \underline{underline}}.}
\end{tabular}}
\end{table*}

\begin{figure*}[!htbp]
\renewcommand{\arraystretch}{0.5}
\setlength\tabcolsep{0.5pt}
\centering
\vspace{-0.2cm}
\resizebox{\textwidth}{!}{ 
\begin{tabular}{ccccccc}
\centering
\includegraphics[width=48mm, height = 36mm]{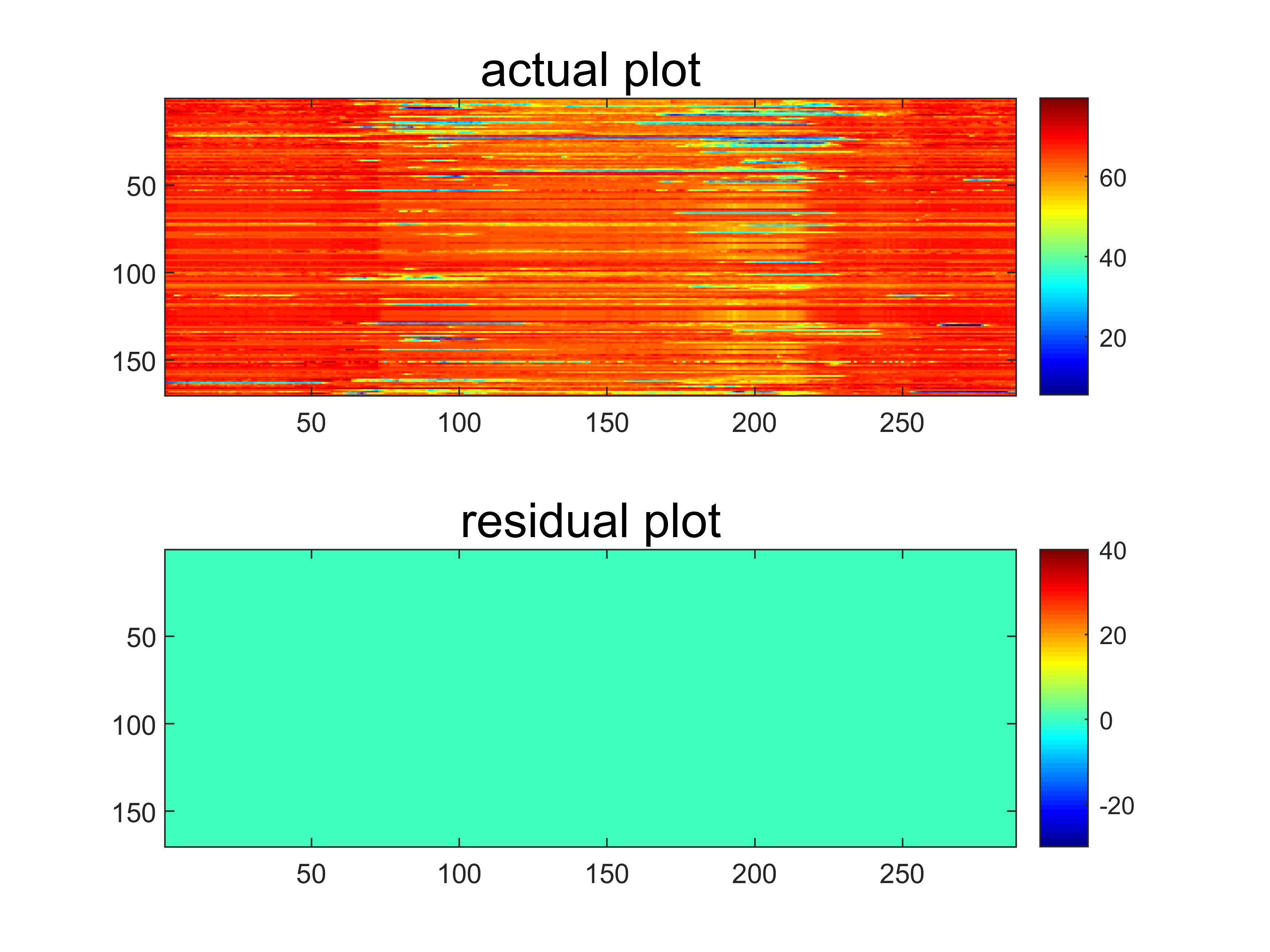}&
\includegraphics[width=48mm, height = 36mm]{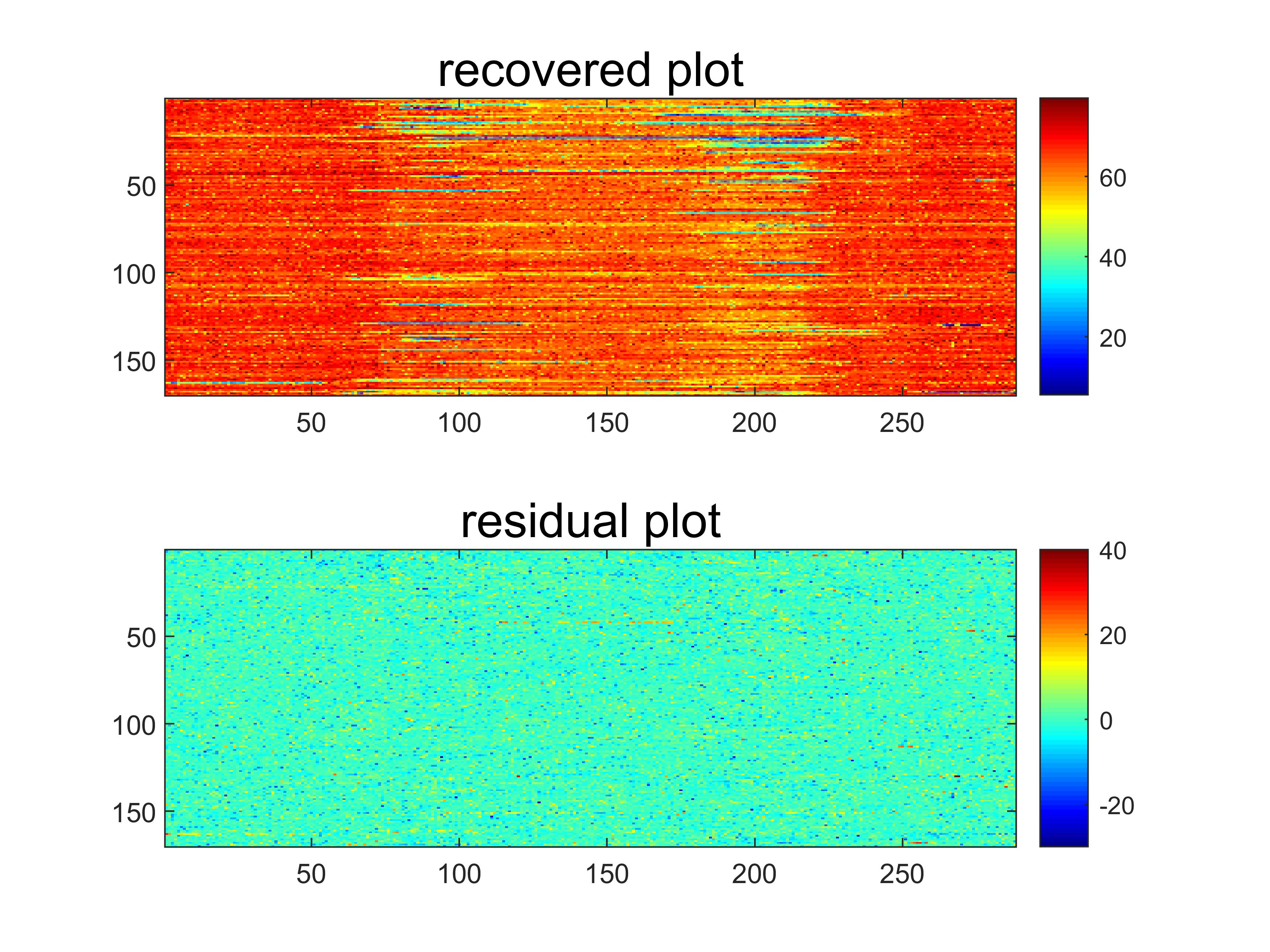}&
\includegraphics[width=48mm, height = 36mm]{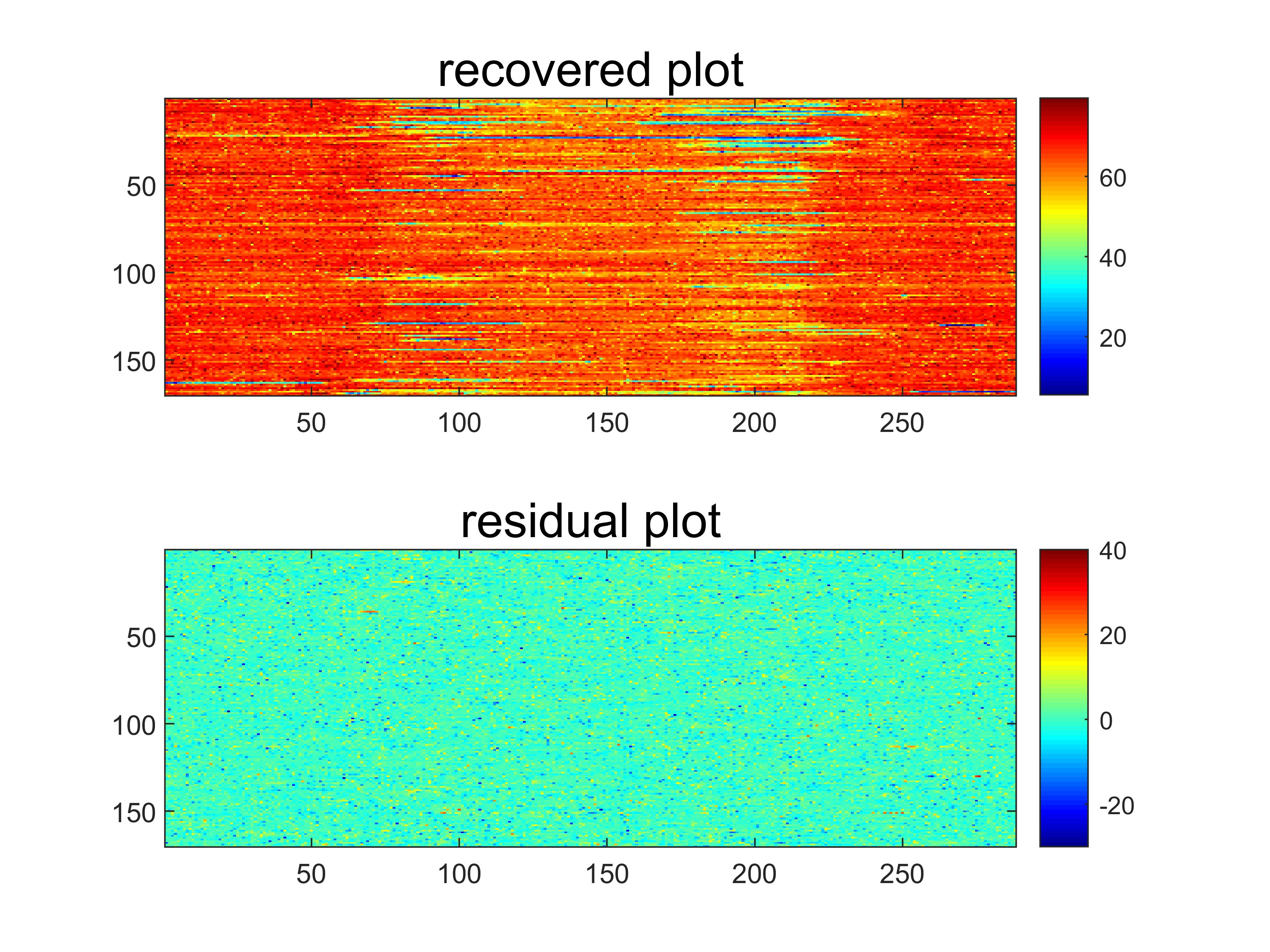}\\
 \scriptsize  \textbf{Ground truth}& \scriptsize \textbf{LRTC-TNN} &\scriptsize \textbf{LATC} \\
\includegraphics[width=48mm, height = 36mm]{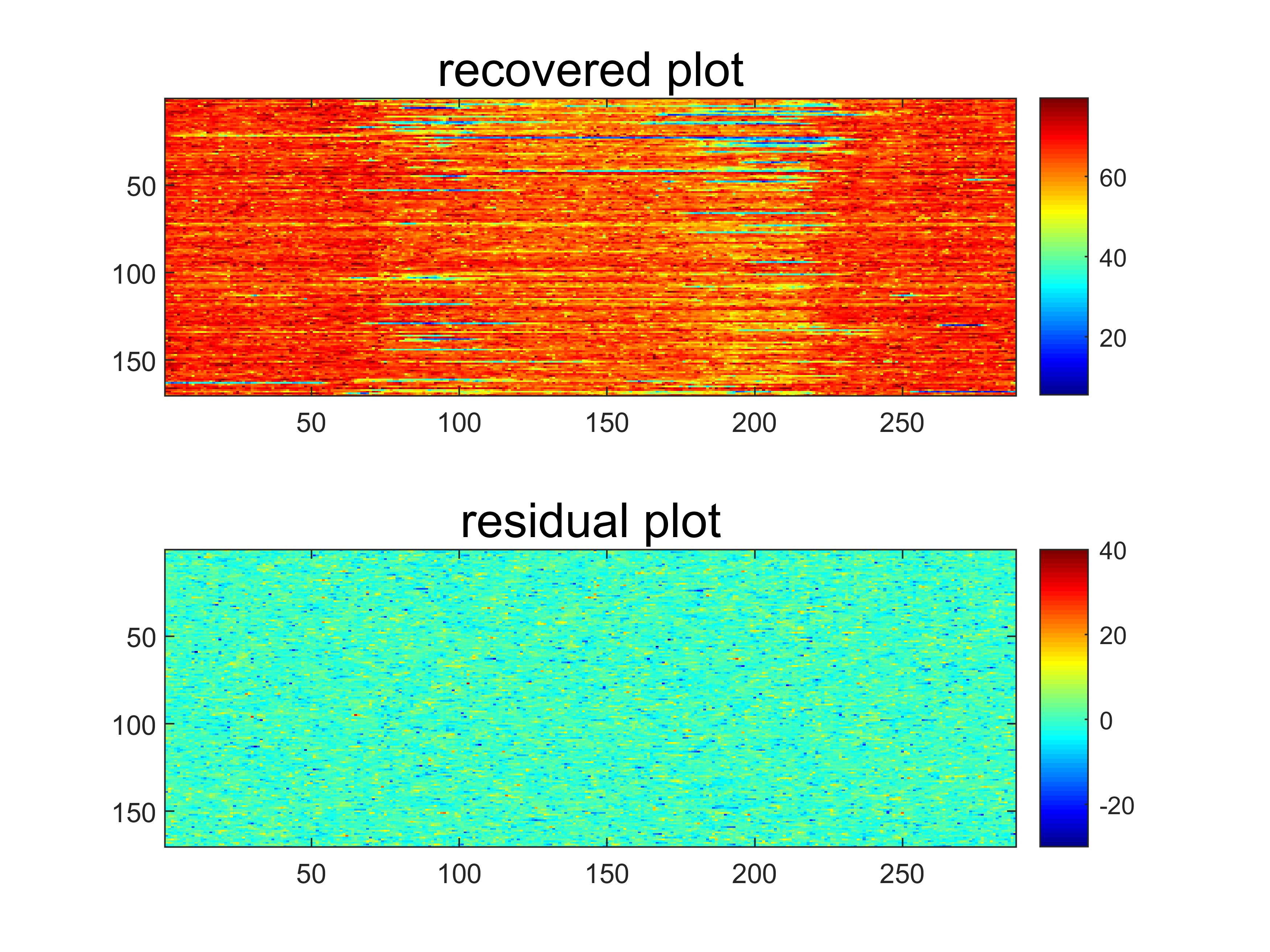}&
\includegraphics[width=48mm, height = 36mm]{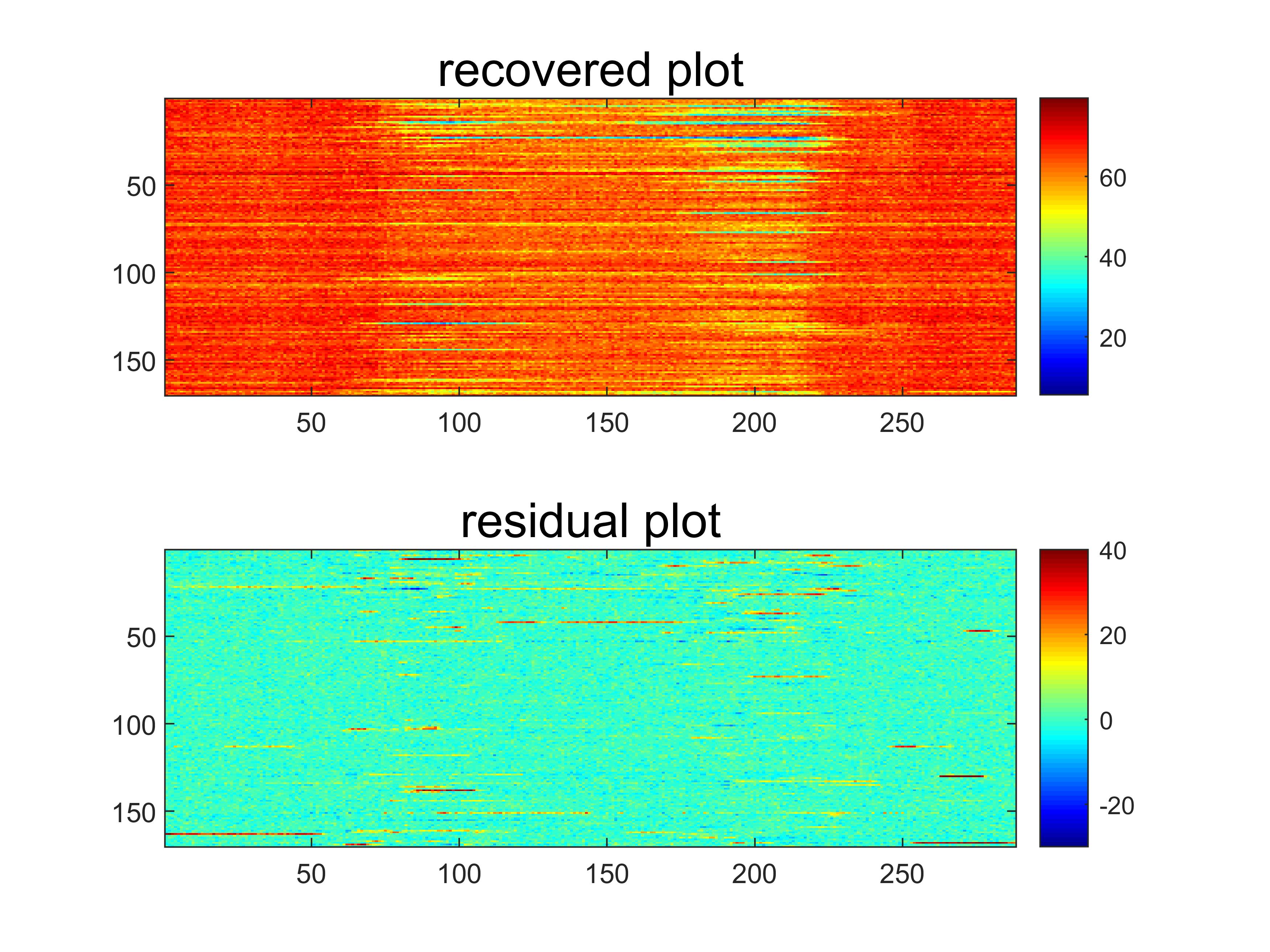}&
\includegraphics[width=48mm, height = 36mm]{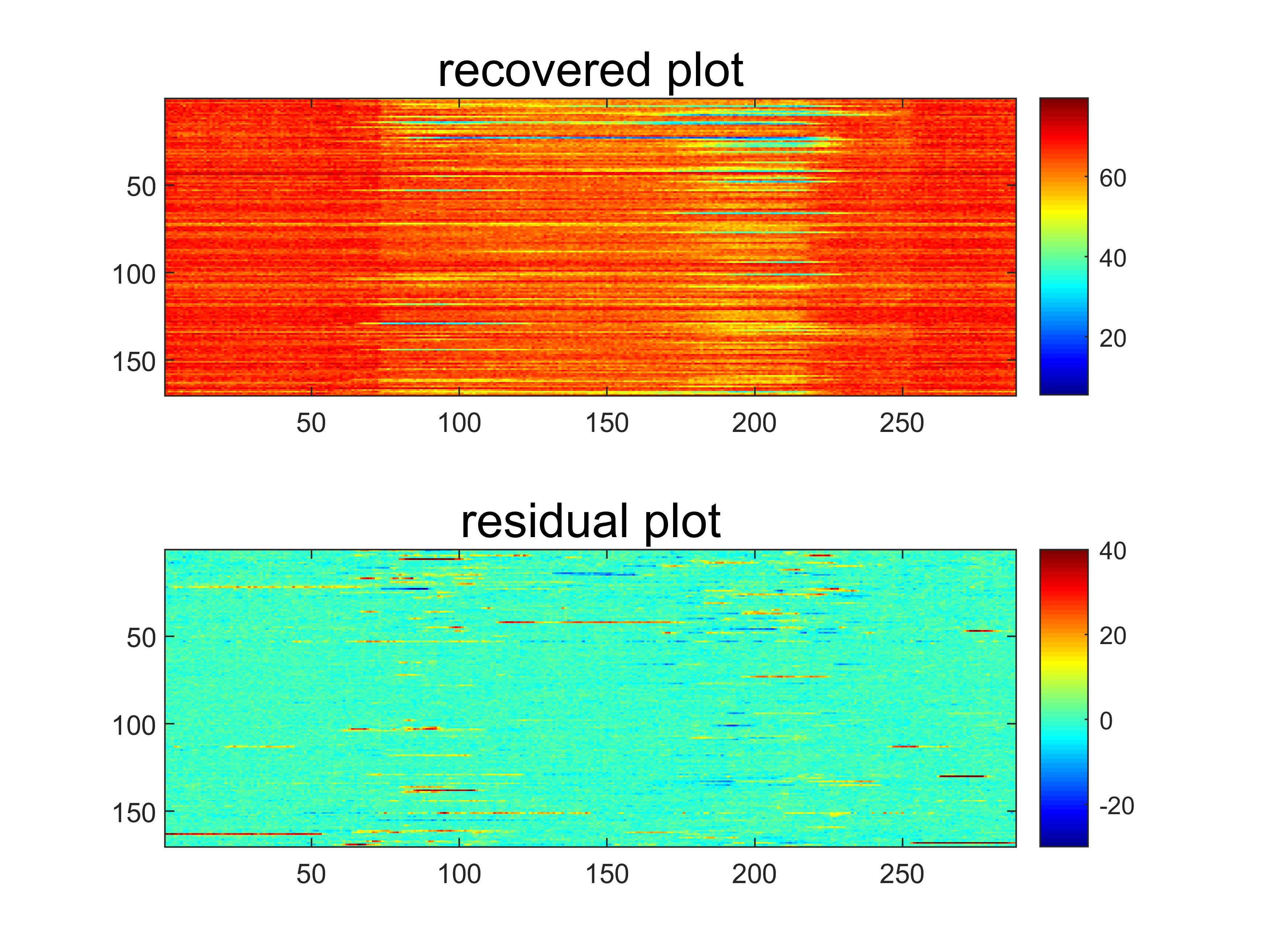}\\
 \scriptsize  \textbf{LRTC-3DST}& \scriptsize \textbf{RTC-SPN} &\scriptsize \textbf{RTC-tubal} \\
 \includegraphics[width=48mm, height = 36mm]{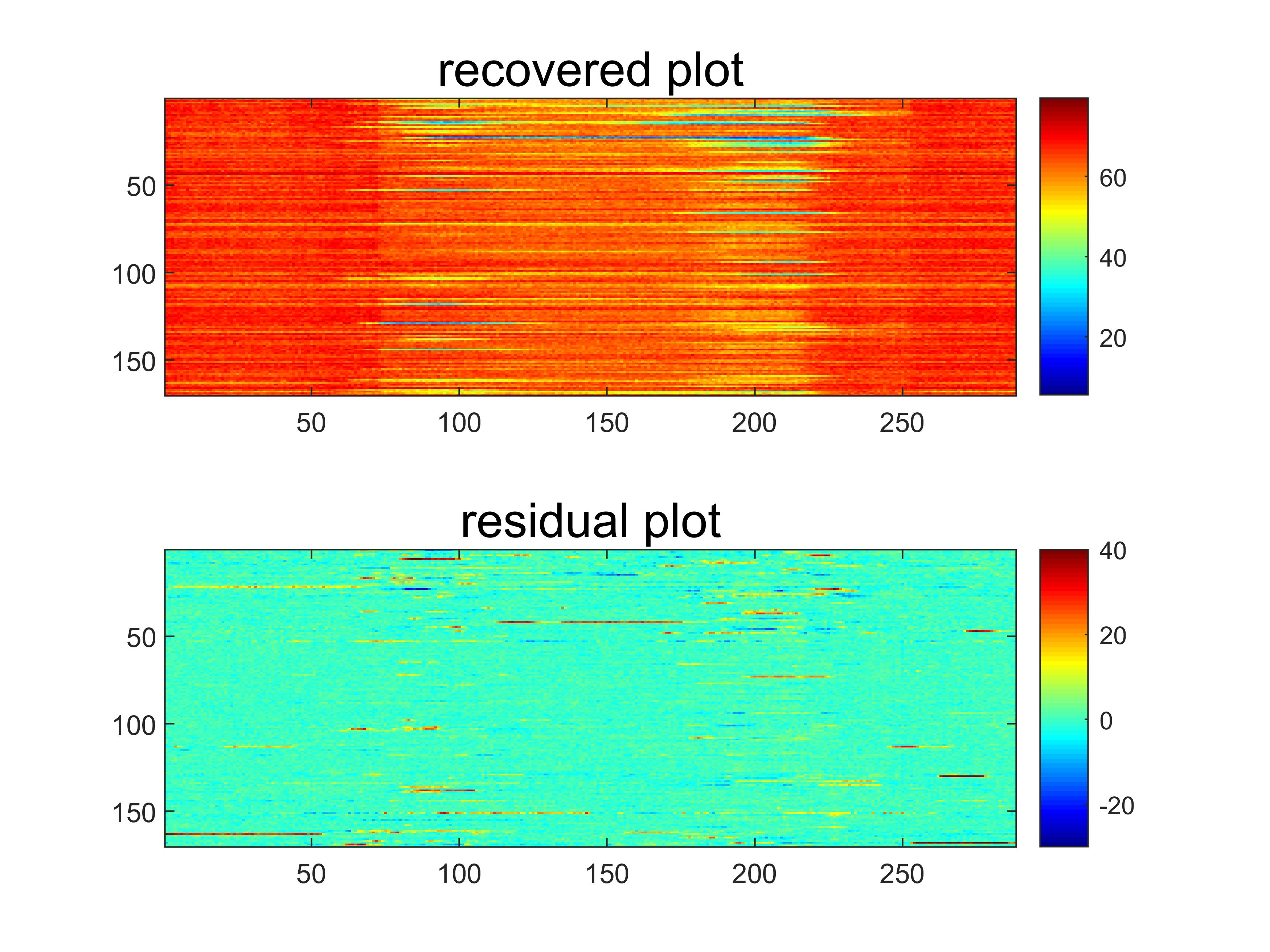}&
\includegraphics[width=48mm, height = 36mm]{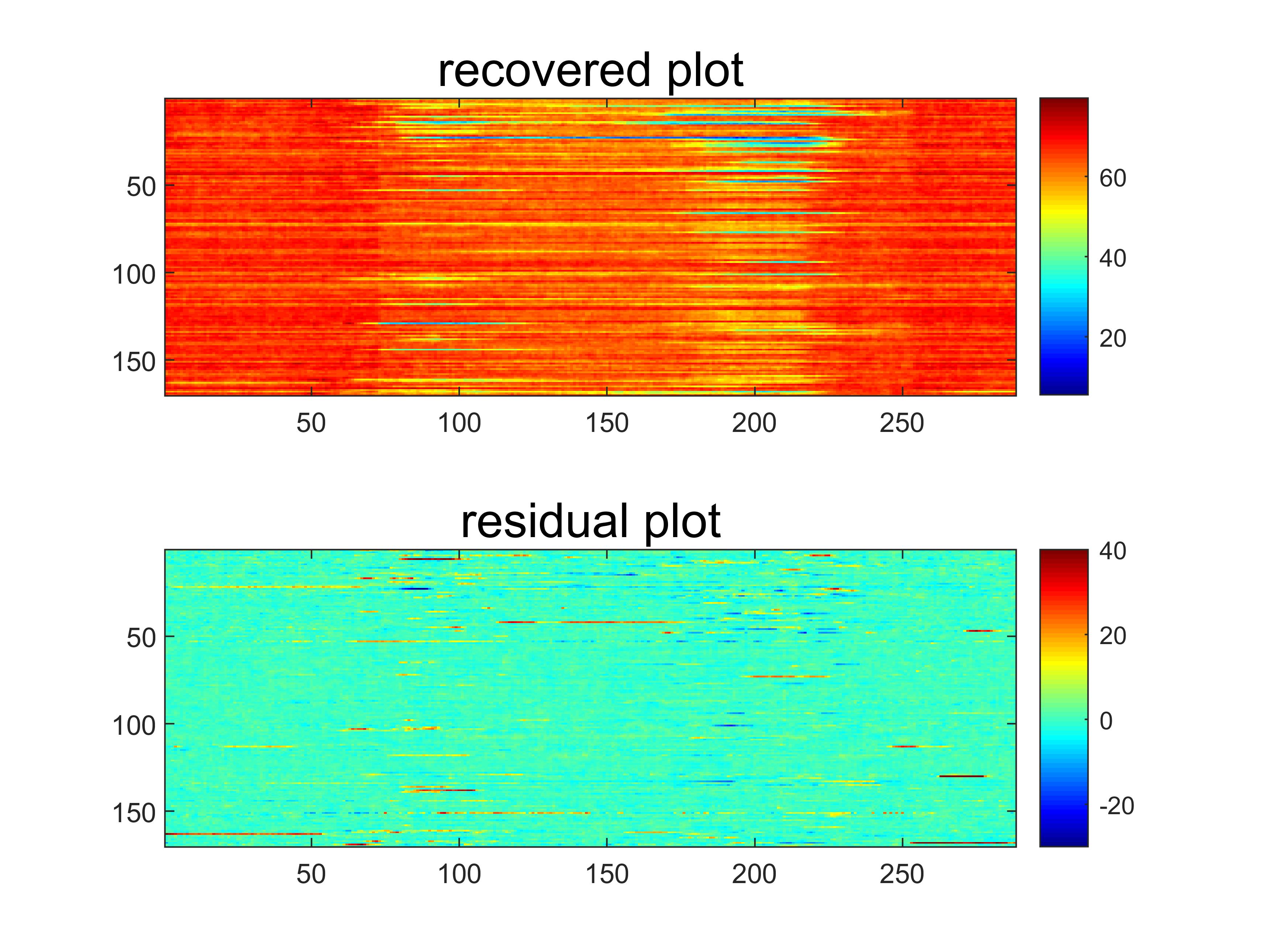}&
\includegraphics[width=48mm, height = 36mm]{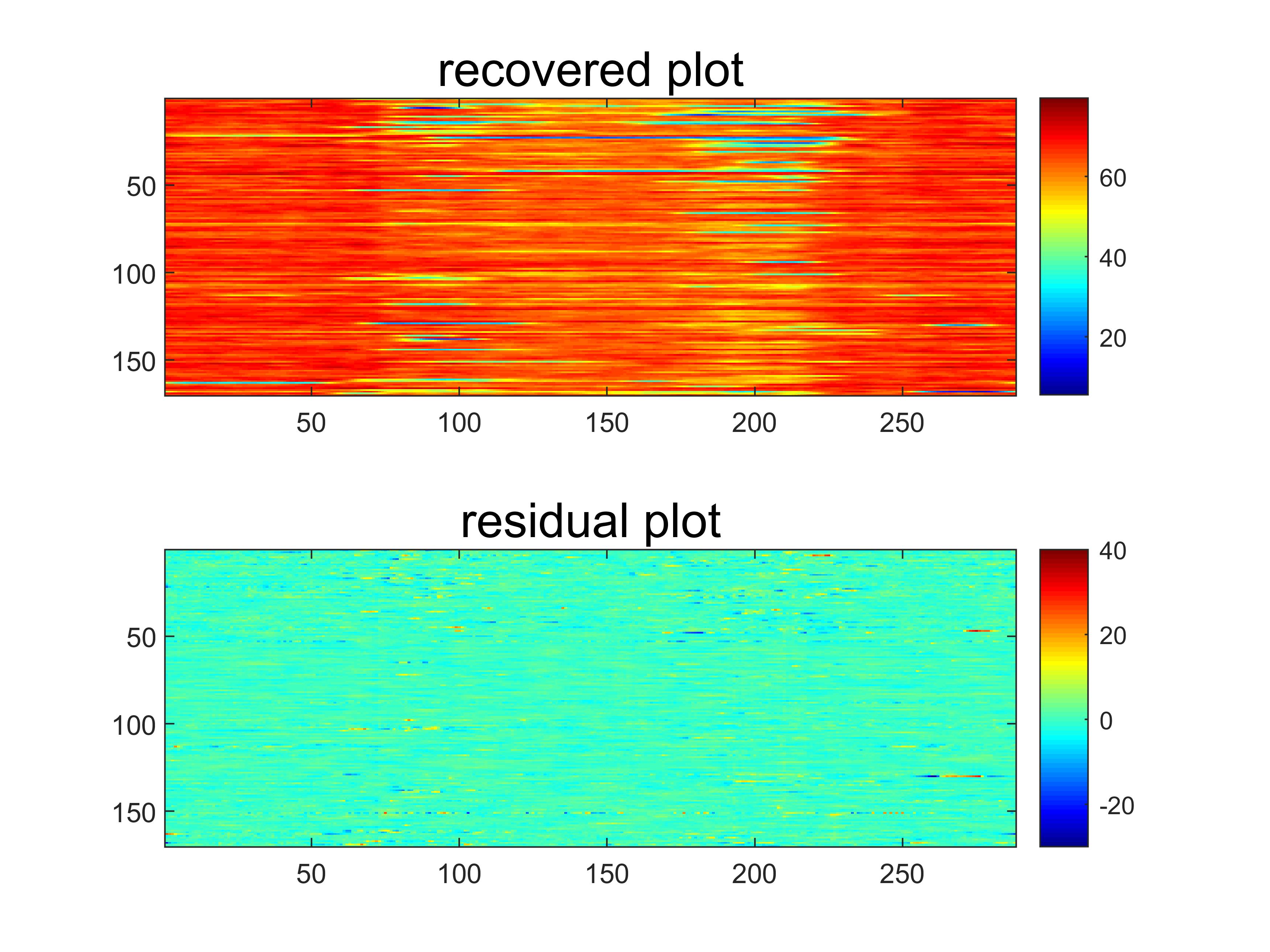}\\
 \scriptsize  \textbf{RTC-TTSVD}& \scriptsize \textbf{RTC-TCTV} &\scriptsize \textbf{RTC-GTNLN} 
\end{tabular}}
\caption{The recovered results (upper) and residual plots (below)  obtained by RTC-GTNLN model and baseline models 
on the 56th day of the PeMS08 dataset. 
}\label{pems08-all-methods}
\end{figure*}

\subsection{Other Discussion}

\begin{figure}[h]
\centering
\includegraphics[width=\linewidth]{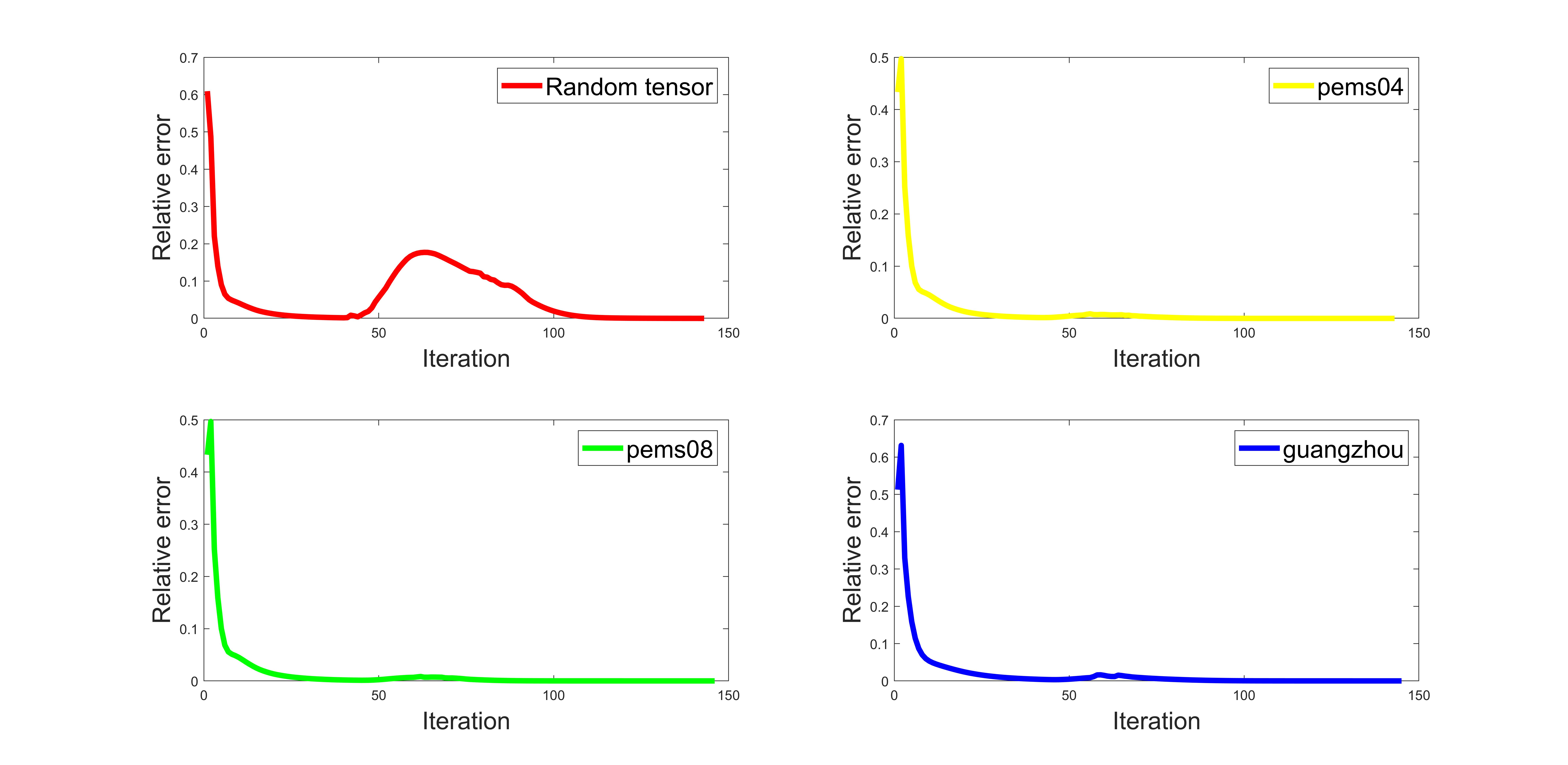}
\caption{Convergence curves of Algorithm 1}
\label{relative_error}
\vspace{-0.4cm}
\end{figure}
\subsubsection{Convergence testing}
To empirically validate the convergence of our proposed Algorithm \ref{alg1}, we construct a synthetic tensor of size 100 × 100 × 100 with entries randomly sampled from the standard normal distribution N(0, 1), augmented with three real-world datasets to form the test tensor. Under the degradation pattern comprising 50\% random missing values and 50\% Laplace noise 1, we monitor the relative error at each iteration using the metric:  
$ {\|\mathbf{X}_{k+1} - \mathbf{X}_k\|_F}/{\|\mathbf{X}_k\|_F}$.
As demonstrated in Figure \ref{relative_error}, the error trajectory reveals stabilized tensor updates after 130 iterations. This  evidence confirms the numerical convergence of our algorithm.

\subsubsection{Ablation study}



The ablation study consists of three key components. 
First, we verify the superiority of the non-convex surrogate TNLN over the traditional convex surrogate.
Second, we demonstrate that GTNLN, as a gradient domain low-rank regularizer, outperforms the original low-rank regularizer in the task of traffic data recovery.
Third, We illustrate that the RTC-GTNLN model that fuses local consistency and global low-rank priors outperforms the separate low-rank and local regularization model.
To ensure comparative consistency across all experiments, we employ a unified robust completion model framework  for ablation studies:


\begin{equation}\label{RTC11}
\begin{aligned}
   \min_{\X,\E}~\mathcal{R} \left( \X \right) &+\lambda\norm{\E}_1,\\
    \text { s.t.}~\Pomega(\X+\E)&= \Pomega(\Y). \\  
    \end{aligned}
\end{equation}

\begin{enumerate}

\item
To demonstrate that our proposed non-convex surrogate TNLN is more effective than the convex surrogate $\norm{\X}_{\circledast}$,
we incorporate $\norm{\X}_{\circledast,\ell}$ and $\norm{\X}_{\circledast}$ as the regularization term $\mathcal{R}(\X)$ in Model (\ref{RTC11}). 
The tests are conducted under various degradation scenarios across multiple datasets. The noise type used is Laplace noise 1, and the missing data pattern followed a random missing scheme. 
As evidenced in Table \ref{tab_nonconvex}, the non-convex surrogate TNLN demonstrates enhanced performance in data recovery tasks compared to the convex Tucker rank surrogate, indicating its superior capacity to capture global low-rank structures.

\begin{table}[t]
\renewcommand{\arraystretch}{1}
\setlength{\tabcolsep}{3pt}
\caption{Performance comparison (in MAE) of proposed non-convex surrogate TNLN and the convex surrogate. }
\centering
\footnotesize
\resizebox{0.35\textwidth}{!}{ 
\begin{tabular}{cc|cccc}
\toprule
Dataset&Degradation cases &  $\norm{\X}_{\circledast}$ & $\norm{\X}_{\circledast,\ell}$\\
\midrule
\multirow{4}{*}{PeMS04}
&20\%missing + 80\%noise & 2.11 & 2.10\\
&40\%missing + 60\%noise & 2.47  & 2.41\\ 
&60\%missing + 40\%noise & 2.84 & 2.72\\
&80\%missing + 20\%noise & 3.89  & 3.55\\ 
\hline
\multirow{4}{*}{PeMS08}
&20\%missing + 80\%noise & 1.83 & 1.81\\
&40\%missing + 60\%noise & 1.95  & 1.90\\ 
&60\%missing + 40\%noise & 2.11 & 2.05\\
&80\%missing + 20\%noise & 2.28  & 2.37\\ 
\hline
\multirow{4}{*}{Guangzhou}
&20\%missing + 80\%noise & 2.97 & 2.87\\
&40\%missing + 60\%noise & 3.32  & 3.18\\ 
&60\%missing + 40\%noise & 3.98 & 3.63\\
&80\%missing + 20\%noise & 6.05  & 4.44\\ 
\midrule
\end{tabular}}
\vspace{-0.3cm}
\label{tab_nonconvex}
\end{table}

\item 
To validate the effectiveness of gradient-domain low-rank regularization, we systematically investigate the performance differences between GTNLN and TNLN by incorporating them into the regularization term $\mathcal{R}(\X)$ of Model (\ref{RTC11}). 
Through comprehensive experiments across multiple degradation scenarios and diverse datasets, employing Laplace noise 1 contamination and random missing patterns, our comparative analysis in Table \ref{tab_grad} reveals that GTNLN demonstrates superior capability in preserving local structural consistency through gradient-domain regularization, and thus has better performance in the data recovery task.

\begin{table}[t]
\renewcommand{\arraystretch}{1}
\setlength{\tabcolsep}{3pt}
\caption{Performance comparison (in RMSE) of proposed  GTNLN and TNLN. }
\centering
\footnotesize
\resizebox{0.37\textwidth}{!}{ 
\begin{tabular}{cc|cccc}
\toprule
Dataset&Degradation cases &  $\norm{\X}_{\circledast,\ell}$ & $\norm{\nabla(\X)}_{\circledast,\ell}$\\
\midrule
\multirow{3}{*}{PeMS04}
&30\%missing + 70\%noise & 4.95 & 2.26\\
&50\%missing + 50\%noise & 5.74  & 2.55\\ 
&70\%missing + 30\%noise & 6.75 & 3.02\\
\hline
\multirow{3}{*}{PeMS08}
&30\%missing + 70\%noise & 4.32 & 2.10\\
&50\%missing + 50\%noise & 4.59  & 2.41\\ 
&70\%missing + 30\%noise & 4.93 & 2.95\\
\hline
\multirow{3}{*}{Guangzhou}
&30\%missing + 70\%noise & 4.89 & 3.03\\
&50\%missing + 50\%noise & 5.38  & 3.35\\ 
&70\%missing + 30\%noise & 6.48 & 3.81\\
\midrule
\end{tabular}}
\vspace{-0.5cm}
\label{tab_grad}
\end{table}


  \item 
To further highlight that the proposed RTC-GTNLN method can encode the global low-rankness and local consistency more efficiently, we introduce the following separated low-rank and local regularization model:
\begin{equation}\label{separate}
\begin{aligned}
    \min_{\X,\E}  ~&\norm{\X}_{\circledast,\ell} + \theta  \normlarge{\nabla(\X)}_F+ \lambda\norm{\E}_1 , \\
    &\text{s.t.} \ \Pomega(\X+\E)= \Pomega(\Y).
    \end{aligned}
\end{equation}
where $\lambda$ is an empirical value determined solely by tensor dimensions that requires no tuning  and $\theta$  is the trade-off parameter that balances global low-rankness and local consistency.
\begin{figure}[t]
\centering
\includegraphics[width=\linewidth]{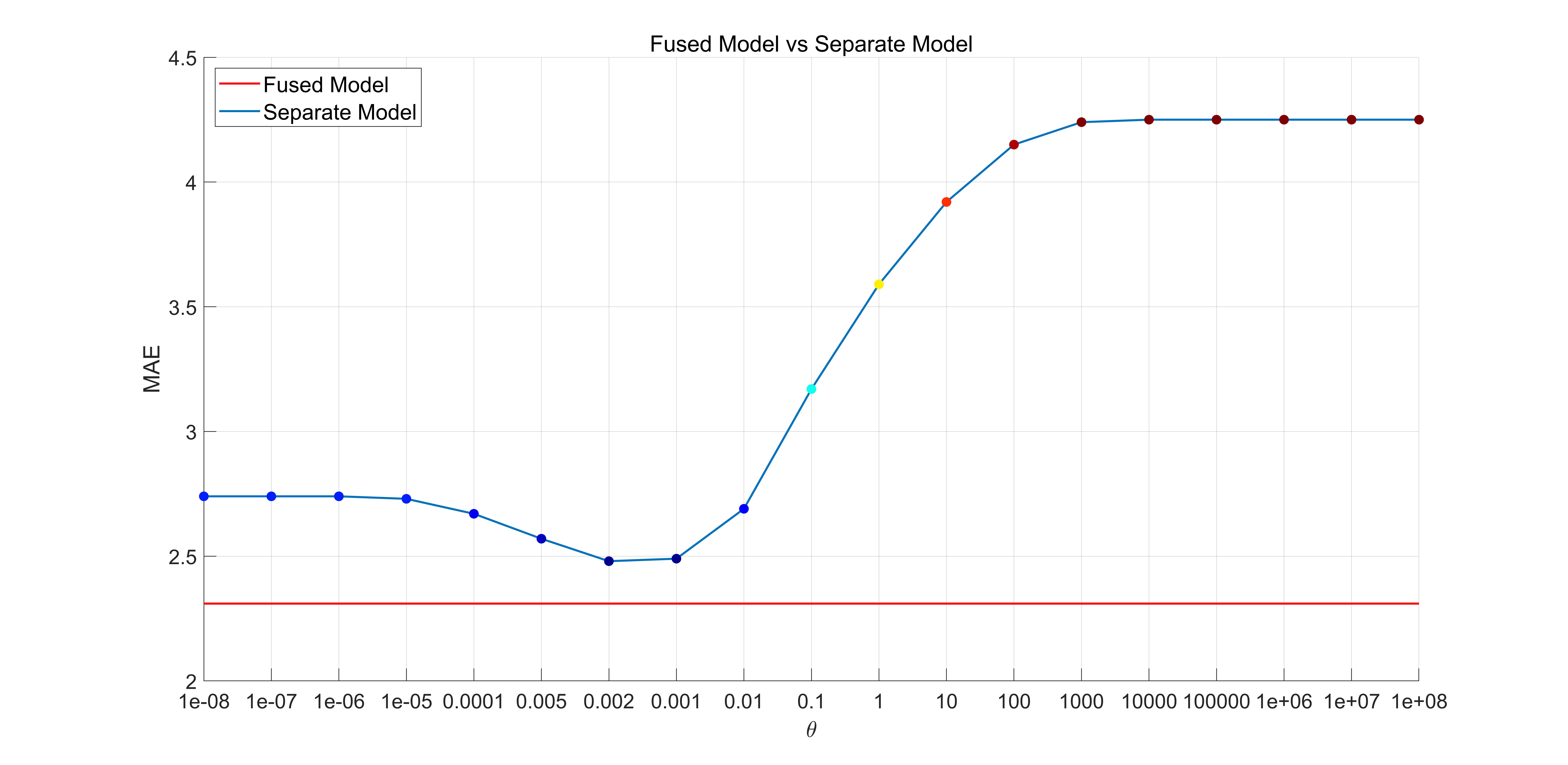}
\caption{Performance comparison (in MAE) of proposed  fused model (\ref{RTC-GTNLN}) and the separated model (\ref{separate}).}
\label{Fused Model vs Separate Model}
\end{figure}
We systematically evaluate the separation model and our proposed fusion model on the Guangzhou dataset under the degradation scenario comprising  50\% random missing values and 50\% Laplace noise 1.
The separation model is tested with 19 parameter configurations of $\theta$ spanning 16 orders of magnitude $(\theta \in [1\times10^{-8}, 1\times10^{8}])$.
As demonstrated in Table 1, our parameter-free model RTC-GTNLN consistently outperformed all parameter-tuned separation models. This empirical evidence reveals that the fusion architecture simultaneously encoding local consistency and global low-rank constraints, achieving two critical advantages:  
1. Parameter-free: Eliminates the need for delicate $\theta$ tuning traditionally required in separation models.  
2. Enhanced recovery: The synergetic regularization mechanism delivers superior reconstruction fidelity compared to standalone parametric approaches.


\end{enumerate}

\section{Conclusion and future directions}

In this study, we propose the RTC-GTNLN model by introducing three effective components: a novel non-convex surrogate TNLN for approximating  tensor Tucker rank, a parameter-free regularizer GTNLN that can simultaneously encode local consistency and global low-rank, and a noise-aware decomposition strategy for separating sparse noise from actual traffic data.
Experimental results demonstrate that the proposed RTC-GTNLN model achieves superior recovery accuracy across various noise and missing data scenarios, surpassing state-of-the-art algorithms in existing literature. However, our current model employs relatively simplistic noise characterization. For practical applications, incorporating Laplacian regularization into the RTC-GTNLN framework could enhance representation of continuous structured noise \citep{hu2024vehicle}. Additionally, given the importance of autocorrelation characteristics in time series data\citep{chen2021low}, developing a unified model that simultaneously captures both autoregressive properties and low-rank structures presents a promising direction for future improvements.

\label{sec:conclusion}



\appendix 
\subsection{Proof of Lemma \ref{GTNLN-TV}}
\begin{proof}
For matrix $X\in \mathbb{R}^{n_1\times n_2}$ with rank $r_0$ and singular value $\sigma_1,\sigma_2,\cdots,\sigma_{r_0}$, it can be verified that
\[(\sqrt{1+\frac{1}{\eta_0(\Xm)}}-1) \norm{\Xm}_{F} \leq \norm{\Xm}_*-\norm{\Xm}_F \leq(\sqrt{r_0}-1)\norm{\Xm}_{F}, \]
$\raisebox{1ex}{where $\eta_0(\Xm)=\operatorname{max}(\frac{\sigma_1(\Xm)}{\sigma_2(\Xm)},\frac{\sigma_2(\Xm)}{\sigma_3(\Xm)},\cdots,\frac{\sigma_{r_0-1}(\Xm)}{\sigma_{r_0}(\Xm)})$. The sec-}$\\ond half of the inequality holds because $\norm{\Xm} \leq\sqrt{r_0}\norm{\Xm}_F$ \citep{recht2010guaranteed}. The first half is established from $\norm{\Xm}_{*}=\sigma_1+\sigma_2+\cdots+\sigma_{r_0}$,
$\norm{\Xm}_{F}=\sqrt{\sigma_1^2+\sigma_2^2+\cdots+\sigma_{r_0}^2}$, and 
$(1+1/\eta_0(\Xm))(\sigma_1^2+\sigma_2^2+\cdots+\sigma_{r_0}^2) \leq (\sigma_1+\sigma_2+\cdots+\sigma_{r_0})^2$.
Therefore, considering the unfold matrix $\Xm_i$ of the third-order tensor $\X$, the following inequality holds true:
\begin{multline}
(\sqrt{1+\frac{1}{\eta_0(\Xm_i)}}-1) \norm{\Xm_i}_{F} \leq \norm{\Xm_i}_*-\norm{\Xm_i}_F\\ \leq(\sqrt{r_i}-1)\norm{\Xm_i}_{F}, i=1,2,3,
\end{multline}
where $r_i=rank(\Xm_i) $ and $\eta_0(\Xm_i)=\operatorname{max}(\frac{\sigma_1(\Xm_i)}{\sigma_2(\Xm_i)},\frac{\sigma_2(\Xm_i)}{\sigma_3(\Xm_i)},\cdots,\\ \frac{\sigma_{r_0-1}(\Xm_i)}{\sigma_{r_0}(\Xm_i)})$. This means
\[(\sqrt{1+\frac{1}{\eta(\X)}}-1) \norm{\X}_{F} \leq \norm{\X}_{\circledast,\ell} \leq(\sqrt{r}-1)\norm{\X}_{F}, \]
where $r=max(rank(\Xm_1),rank(\Xm_2),rank(\Xm_3))$, $\eta(\X)=max(\eta_0(\Xm_1),\eta_0(\Xm_2),\eta_0(\Xm_3))$. Replace $\X$ with its gradient tensor $\G=\nabla(\X)$, then
\[(\sqrt{1+\frac{1}{\eta(\G)}}-1) \norm{\nabla(\X)}_{F} \leq \norm{\nabla(\X)}_{\circledast,\ell} \leq(\sqrt{r}-1)\norm{\nabla(\X)}_{F}, \]
where $r=max(rank(\Gm_1),rank(\Gm_2),rank(\Gm_3))$, $\eta(\G)=max(\eta_0(\Gm_1),\eta_0(\Gm_2),\eta_0(\Gm_3))$. 
Finally, the lemma \ref{GTNLN-TV} holds due to $\norm{\X}_{TV} = \norm{\nabla(\X)}_F$.
\end{proof}



\ifCLASSOPTIONcaptionsoff
  \newpage
\fi



 \bibliographystyle{IEEEtranN}
 \bibliography{main}






\end{document}